\def\year{2022}\relax
\documentclass[letterpaper]{article} 
\usepackage{aaai22}  
\usepackage{times}  
\usepackage{helvet}  
\usepackage{courier}  
\usepackage[hyphens]{url}  
\usepackage{graphicx} 
\usepackage{natbib}  
\usepackage{caption} 
\DeclareCaptionStyle{ruled}{labelfont=normalfont,labelsep=colon,strut=off} 
\title{Trade-offs between membership privacy \& adversarially robust learning}
\author{
    Jamie Hayes
}
\affiliations{
   University College London\\
    \texttt{j.hayes@cs.ucl.ac.uk}
}

\usepackage[utf8]{inputenc} 
\usepackage[T1]{fontenc}    
\usepackage{hyperref}       
\usepackage{url}            
\usepackage{booktabs}       
\usepackage{amsfonts}       
\usepackage{nicefrac}       
\usepackage{microtype}      

\usepackage{bm}
\usepackage{url}
\usepackage{float}
\usepackage{amsmath}
\usepackage{amssymb}
\usepackage{amsthm}
\usepackage{xspace}
\usepackage{color}
\usepackage{graphicx}
\usepackage{grffile}
\usepackage{soul}
\usepackage{etoolbox}
\usepackage{chngpage}
\usepackage{environ}
\usepackage{subcaption}
\usepackage{rotating}
\usepackage{multirow}
\usepackage{algorithm}
\usepackage[noend]{algpseudocode}
\usepackage{hyperref}
\usepackage{cleveref}
\usepackage{bigints}
\usepackage{mathtools}
\usepackage{wrapfig}
\usepackage{natbib}

\def\eqref#1{equation~\ref{#1}}

\def\1{\bm{1}}

\DeclareMathAlphabet{\mathsfit}{\encodingdefault}{\sfdefault}{m}{sl}
\SetMathAlphabet{\mathsfit}{bold}{\encodingdefault}{\sfdefault}{bx}{n}

\DeclareMathOperator*{\argmin}{arg\,min}

\DeclarePairedDelimiter{\norm}{\lVert}{\rVert}

\DeclareMathOperator{\sign}{sign}

\DeclarePairedDelimiter\abs{\lvert}{\rvert}%
\DeclarePairedDelimiterX{\inp}[2]{\langle}{\rangle}{#1, #2}

\makeatletter
\newtheorem*{rep@theorem}{\rep@title}
\newcommand{\newreptheorem}[2]{%
\newenvironment{rep#1}[1]{%
 \def\rep@title{#2 \ref{##1}}%
 \begin{rep@theorem}}%
 {\end{rep@theorem}}}
\makeatother
\newtoggle{shortpaper}

\newtheorem{definition}{Definition}[section]

\newtheorem{prop}{Proposition}

\newreptheorem{lemma}{Lemma}
\newreptheorem{theorem}{Theorem}
\newreptheorem{conjecture}{Conjecture}
\newreptheorem{statement}{Statement}
\newreptheorem{definition}{Definition}
\newreptheorem{corollary}{Corollary}
\newreptheorem{prop}{Proposition}
\newreptheorem{example}{Example}

\begin{document}

\maketitle

\begin{abstract}

Historically, machine learning methods have not been designed with security in mind.
In turn, this has given rise to adversarial examples, carefully perturbed input samples aimed to mislead detection at test time, which have been applied to attack spam and malware classification~\citep{dalvi2004adversarial,lowd2005adversarial, lowd2005good}, and more recently to attack image classification~\citep{szegedy2013intriguing}. 
Consequently, an abundance of research has been devoted to designing machine learning methods that are robust to adversarial examples. 
Unfortunately, there are desiderata besides robustness that a secure and safe machine learning model must satisfy, such as fairness and privacy. 
Recent work by~\citet{song2019privacy} has shown, empirically, that there exists a trade-off between robust and private machine learning models. 
Models designed to be robust to adversarial examples often overfit on training data to a larger extent than standard (non-robust) models. 
If a dataset contains private information, then any statistical test that separates training and test data by observing a model's outputs can represent a privacy breach, and if a model overfits on training data, these statistical tests become easier. 
The inversely proportional relationship between privacy and overfitting has been well documented in previous works~\citep{yeom2018privacy, rice2020overfitting}. 

In this work, we identify settings where standard models will overfit to a larger extent in comparison to robust models, and as empirically observed in previous works, settings where the opposite behavior occurs. 
Thus, it is not necessarily the case that privacy must be sacrificed to achieve robustness. 
The degree of overfitting naturally depends on the amount of data available for training. 
We go on to characterize how the training set size factors into the privacy risks exposed by training a robust model on a simple Gaussian data task, and show empirically that our findings hold on image classification benchmark datasets, such as CIFAR-10 and CIFAR-100.

\end{abstract}

\section{Introduction}
\label{sec:introduction}

\begin{figure}[t]
\centering
  \includegraphics[width=0.7\linewidth]{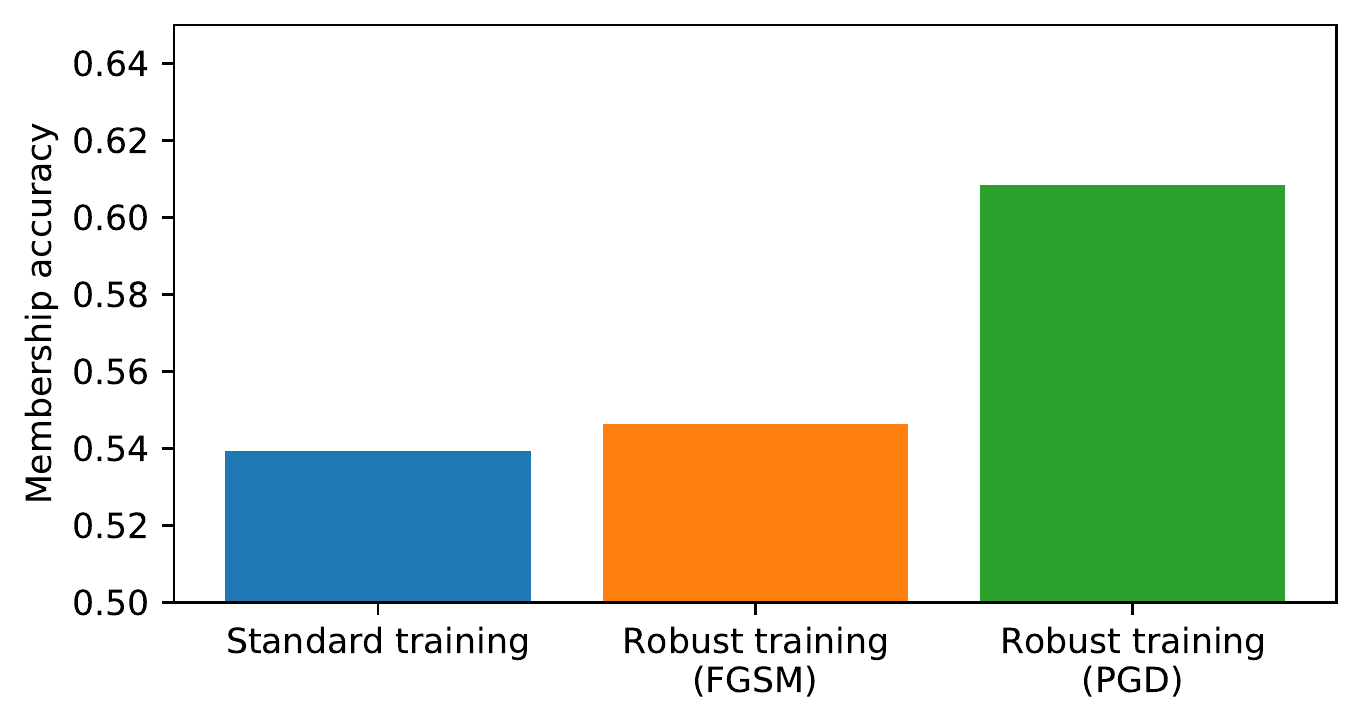}
\caption{Membership accuracy of standard and robust classifiers on the CIFAR-10 dataset. We measure the accuracy in determining if an input belongs to the test set or a subset of the training set that is of equal size. The baseline membership accuracy is therefore 50\%.}
\label{fig:cifar10_mem_inf_full_train}
\end{figure}

Overfitting is the enemy of generalization -- a fundamental property of any useful machine learning model. 
If a model overfits on its training data, it is invariably more confident in its predictions on these inputs.
Statistical tests can then be used to discriminate between a models behavior on the training set and test set, and if inclusion of an input in the training set is in some way sensitive, this can cause a privacy violation. 
Recently,~\citet{song2019privacy} have shown that securing models against adversarial examples~\citep{szegedy2013intriguing, biggio2018wild} can exacerbate this privacy issue. 
This is demonstrated in~\cref{fig:cifar10_mem_inf_full_train}: we train a ResNet-18 classifier~\citep{he2016deep} on the CIFAR-10 dataset~\citep{krizhevsky2009learning} using standard gradient descent (resulting in a \emph{standard model}), and via adversarial training (resulting in a \emph{robust model}) using either the FGSM~\citep{goodfellow2014explaining} or PGD~\citep{madry2017towards} attack (see \cref{sec:cifar10_results} for full details).
We then exploit overconfidence in the model's predictions to determine if an input belongs to the training set or the test set, where we select a random subset of the CIFAR-10 training set that is equal in size to the test set to determine membership. 

We plot the \emph{membership accuracy} -- the accuracy of determining if an input belongs to the training set or test set, and note that this is always higher on an adversarially robust trained model in comparison to a standard model. 
We note that the results in~\cref{fig:cifar10_mem_inf_full_train} are not state-of-the-art -- the exact attack method used here to determine membership is largely unimportant, what \emph{is important} is the  disparate vulnerability to membership attacks between a standard model and an adversarially robust model. 

Following work by~\cite{sablayrolles2019white}, we show that vulnerability to membership attacks is proportional to the expected gap between training and test loss, which we refer to as the \emph{loss gap}. In other words, the loss gap corresponds to how much the model overfits.
In contrast to empirical work~\citep{song2019privacy}, that shows overfitting is almost always exacerbated by adversarial training, we prove in a simple data setting, adversarial training can either increase or decrease the risk of membership attacks, because the expected loss gap can increase or decrease. 
We show that the precise cases where adversarially robust training has a smaller or larger loss gap in comparison to standard training, depends on the size of perturbation used in adversarial training, $\epsilon$, and the size of the training set.
After presenting this characterization in a simple data setting, we empirically demonstrate this privacy-robustness trade-off (or lack thereof) exists in more complex datasets, such as CIFAR-10.

In sum, we make the following contributions:

\begin{enumerate}
    \item We prove that robust models (through adversarial training) can be \emph{more} private than standard models. Additionally, we present settings where the opposite behavior occurs; robust models are provably less private than their standard counterparts.
    \item We show that vulnerability to membership attacks depends on the size of the perturbation used in adversarial training \emph{and} the size of the training set.
    \item We empirically show these relationships hold on complex image datasets such as CIFAR-10.
\end{enumerate}

\section{Background}
\label{sec:background}

The most relevant works to our own come from~\citet{song2019privacy} and~\citet{rice2020overfitting}.
\citet{song2019privacy} led an empirical evaluation of the privacy risks that are exacerbated by training models to be robust to adversarial examples. 
Firstly, they introduce a new membership attack that delineates between training and test inputs based on the \emph{adversarial} loss (the loss on adversarial examples) rather than standard loss. 
They then show membership accuracy is higher on robust models than on standard models, and this is true regardless of if the attack uses the standard or adversarial loss to predict membership. 
The specific method used to create a robust model does not reduce the disparate membership vulnerability between robust and standard models; \citet{song2019privacy} show an increase in membership attack accuracy over a standard model, when the robust model is constructed using adversarial training~\citep{madry2017towards}, or using a certified adversarial example defense that guarantees robustness around an $\epsilon$-ball on training inputs~\citep{gowal2018effectiveness, mirman2018differentiable, wong2017provable}. 
The intuition as to why robust models are more susceptible to privacy attacks goes as follows -- robust models are trained to be insensitive to perturbations around their training inputs, but by doing so, they are made \emph{more} sensitive to the choice of training inputs, because the invariance to perturbations within an $\epsilon$-ball around a training input does not completely generalize to test inputs. 
Inspired by \citet{song2019privacy}, our work seeks to answer the following questions: 
Is it always the case that robust models reduce privacy, or are there cases where robust models are also more private?
How does the training set size factor into the level of privacy of a model?

\citet{rice2020overfitting} show that \emph{robust} overfitting is a concern in robust models. 
The gap between standard train and test error is smaller than the gap between robust train and test error -- where we define robust error as the worst-case error within an $\epsilon$-ball around the inputs. 
However, they show that robust overfitting can be mitigated by simple defenses such as early stopping.
\citet{yeom2018privacy} formally analyze the connection between overfitting and privacy, and find that overfitting is sufficient to allow an attacker to perform membership attacks, and additionally, attribute inference attacks~\citep{fredrikson2014privacy, fredrikson2015model, wu2016methodology}.  
In light of work by~\cite{yeom2018privacy}, our findings on the connection between membership attacks and overfitting is not entirely novel, but we transcribe this connection into the membership attack language introduced by~\citet{sablayrolles2019white}, who propose a Bayes optimal membership attack and demonstrate state-of-the-art results on common image dataset benchmarks.

The privacy risks of overfitting have also been exposed by
\citet{carlini2019secret}, where they show that it is possible to extract sensitive information from the training set simply by querying the trained model. Using an attack based on log-perplexity of a model's output, they show it is possible to extract secrets in language based models such as credit card information.
\citet{panprivacy} and \citet{song2020information} have recently investigated the privacy risks of memorization in powerful language models such as GPT-2~\citep{radford2019language} and BERT~\citep{devlin2018bert}, finding that text embeddings from these models capture a lot sensitive information about the plain text inputs.

To the best of our knowledge, the work of~\citet{homer2008resolving} was the first to highlight a privacy risk from inferring general dataset statistics. Specifically, they infer the presence of genomes within a dataset by comparing with published statistics about this dataset, and this was further studied in~\citet{dwork2015robust} and~\citet{backes2016membership}. 
As far we are aware, the first work to specifically target \emph{membership inference} of single inputs was~\citet{shokri2017membership}. 
By training shadow models that learn the distribution of outputs of a target model, they show it is possible to infer membership of inputs on common image datasets (MNIST~\citep{lecun1998gradient} and CIFAR-10~\citep{krizhevsky2009learning}) and tabular data (ADULT dataset~\citep{kohavi1996scaling}).
\citet{sablayrolles2019white} then went on to show, under limited assumptions, black-box membership attacks are equivalent to white-box membership attacks. 
That is, they show that the Bayes optimal membership attack only inspects the loss of an input, and so devising a membership attack that can inspect internal parameters of a model should perform no  better.
However, in practice, white-box membership attacks have been shown by~\citet{nasr2018comprehensive} to outperform black-box membership attacks.
Differential privacy~\citep{dwork2006calibrating} has been touted as a potential remedy to privacy attacks~\citep{carlini2019secret}. However, recent work by~\citet{bagdasaryan2019differential} has shown that differentially private models have their own problems -- differential privacy often increases average test error, however poorly represented subpopulations in the data distribution incur a much larger increase in test error, which in turn cases fairness concerns.

Although many previous works have investigated the relationship between generalization error and robustness, the axis of interest in these works is usually adversarial robustness of the final model and the trade-off with standard generalization error. 
This is not a primary axis of interest in our work, we are concerned with how robust training relates to overfitting (and thus privacy) as a function of both the size of available training data \emph{and} the size of $\epsilon$ used in robust training.
In an orthogonal direction to our work on privacy, \citep{raghunathan2019adversarial, raghunathan2020understanding, chen2020more, min2020curious, nakkiran2019adversarial, schmidt2018adversarially, tsipras2018robustness, tsipras2018there, zhang2019theoretically, dohmatob2018generalized, carmon2019unlabeled, najafi2019robustness, uesato2018adversarial} have all studied trade-offs between generalization error and robustness to adversarial examples.
\section{Membership inference and overfitting}
\label{sec:membership_inference_and_overfitting}

We follow the same set-up as introduced by~\cite{sablayrolles2019white}.
Let $\mathcal{X}\times\mathcal{Y}$ be a data distribution, from which we sample $n\in\mathbb{N}$ training points $(x_1,y_1),\dots,(x_n,y_n)\in\mathcal{X}\times\mathcal{Y}$. 
Given a machine learning model, $f_{\theta}$, a training procedure selects parameters, $\theta$, solving $\argmin_{\theta} \frac{1}{n}\sum^n_{i=1}\ell(f_{\theta}(x_i), y_i)$, where $\ell(f_{\theta}(\cdot), \cdot)$ is a loss function that incurs a large cost when $f_{\theta}(x)\neq y$, and a small cost when $f_{\theta}(x)= y$. 
We assume the posterior distribution follows

\begin{align}
    P\big(f_{\theta}\mid(x_1,y_1),\dots,(x_n,y_n)\big) \propto e^{-\sum^n_{i=1}\ell(f_{\theta}(x_i), y_i)} \label{eq:posterior_dist}
\end{align}

where the randomness either comes from the training procedure (Bayesian methods), or arises due to stochasticity in data sampling.
Given a machine learning model, $f_{\theta}$, a membership inference attack attempts to exploit memorization within $\theta$ to infer if an input $x\in\mathcal{X}$ belongs to the training set.
\cite{sablayrolles2019white} formalizes this by defining binary membership variables, $m_1,\dots,m_n$, where $m_i=0$ for test inputs, $m_i=1$ for training inputs, and the probability of membership is fixed, $P(m_i=1)=\lambda$. 
Then,~\cref{eq:posterior_dist} becomes

\begin{align}
    P\big(f_{\theta}\mid(x_1, y_1,m_1),\dots,(x_n,y_n,m_n)\big) \propto e^{-\sum^n_{i=1}m_i\ell(f_{\theta}(x_i), y_i)} \label{eq:posterior_dist_with_membership}
\end{align}

Formally, a membership attack on a sample, $(x, y)$, aims to compute the following:

\begin{definition}
\label{def:membership_inference}
\citep{sablayrolles2019white} Membership inference of $(x,y)\in\mathcal{X}\times\mathcal{Y}$ amounts to computing:
\begin{align}
    \mathcal{M}(f_{\theta}, x, y) \coloneqq P\big(m=1\mid f_{\theta}, x, y\big) \label{eq:membership_defn}
\end{align}
\end{definition}

Under \cref{def:membership_inference}, \cite{sablayrolles2019white} proved that:

\begin{align}
    \mathcal{M}(f_{\theta}, x, y) \propto \tau(x, y) - \ell(f_{\theta}, x, y) \label{eq:sab_mem_inf_theorem}
\end{align}

where $\tau(x, y) = -\log\big(\int_{\omega}e^{-\ell(f_{\omega}, x, y)}p(f_{\omega})d\omega\big)$, and can be viewed as a calibration threshold, to which the loss is compared in an attack. Throughout this work we associate privacy with resistance to membership inference of training data. \Cref{def:membership_inference} implies if a model does not leak private information then \cref{eq:membership_defn}=$\nicefrac{1}{2}$, providing equally sized training and test sets.

On general learning tasks, this assumption may be particularly onerous. However, we argue in the specific settings we discuss in this work -- Gaussian data and simple image classification -- this assumption is entirely reasonable. For the Gaussian model, it is extremely unlikely one data point dominates in determining the decision boundary, and for image classification problems such as MNIST, CIFAR-10 and ImageNet, the difference in test accuracy when removing the most influential data point is almost identical to if it had not been removed -- see Figure 2 in \citet{feldman2020neural} where the test accuracy when training on 99.9\% of training set is identical to training on the full training set (where the removed 0.1\% are the training points that are determined to be most likely to be memorised and so exhibit strong membership identifiability). Furthermore, the assumption could in fact be somewhat relaxed to instead assume $P(f_\theta | m_1 = 1, x_1, y_1)$ is within a constant factor $c$ of $P(f_\theta | m_2 = 1, x_1, y_1)$, and the results and proofs would be identical upto a constant factor.

The normalizing constant in \cref{eq:membership_defn} will decompose like $\lambda P\big(f_{\theta}\mid m_i=1, x_i, y_i\big) + (1-\lambda)P\big(f_{\theta}\mid m_i=0, x_i, y_i\big)$. 
It is possible that $P\big(f_{\theta}\mid m_1=1, x_1, y_1\big) \neq P\big(f_{\theta}\mid m_2=1, x_2, y_2\big)$, however we argue that it is a reasonable to assume $P\big(f_{\theta}\mid m_1=1, x_1, y_1\big) \approx P\big(f_{\theta}\mid m_2=1, x_2, y_2\big)$, where $(x_1, y_1)$ and $(x_2, y_2)$ are arbitrary draws from the data distribution, since in standard image classification tasks (e.g. MNIST, CIFAR-10, ImageNet) the variance of influence of a single training example on the final model is relatively small \citep{koh2017understanding, basu2020influence}. 
Under this mild assumption, we can define a measure of how much $(x_1,y_1)$ leaks about membership in comparison to a reference input, $(x_2,y_2)$:

\begin{definition}
\label{def:comparitive_membership_inference}
Measuring how much $(x_1,y_1)$ leaks about membership in comparison to $(x_2,y_2)$ amounts to computing:
\begin{align}
\begin{split}
    \mathcal{M}_{c}(f_{\theta}, x_1, y_1, x_2, y_2) \coloneqq& P\big(m_1=1\mid f_{\theta}, x_1, y_1\big) \\
    -& P\big(m_2=1\mid f_{\theta}, x_2, y_2\big) \label{eq:comparitive_membership_defn}
\end{split}
\end{align}
\end{definition}

Clearly if $\abs{\mathcal{M}_{c}}=1$,  membership information is leaked by only one input, and $\mathcal{M}_{c} = 0$ implies the membership information leaked by $(x_1, y_1)$ is identical to the membership information leaked by $(x_2, y_2)$. 
From \cref{eq:sab_mem_inf_theorem}, we have the following relation:

\begin{align}
\begin{split}
    \mathcal{M}_{c}(f_{\theta}, x_1, y_1, x_2, y_2) \propto& \ell(f_{\theta}, x_2, y_2) - \tau(x_2, y_2) \\
    +& \tau(x_1, y_1) - \ell(f_{\theta}, x_1, y_1)  \label{eq:comp_mem_inf_relation}
\end{split}
\end{align}

Let $\mathcal{D}^{\text{tr}}, \mathcal{D}^{\text{te}} \subseteq \mathcal{X}\times\mathcal{Y}$ denote training and test sets of equal size $n\in \mathbb{N}$, respectively. 
Furthermore, let $(x,y)\in\mathcal{D}^{\text{tr}}$ and $(x^*,y^*)\in\mathcal{D}^{\text{te}}$, with associated membership variables $m$ and $m^*$ satisfying 
$P(m = 1)$ and $P(m^* = 0)$.
We can measure the average comparative membership information that can inferred from a training set input in comparison to a reference test set input by computing the following:

\begin{align}
    \mathop{\mathbb{E}}_{\substack{(x,y)\in\mathcal{D}^{\text{tr}}\\ (x^*,y^*)\in\mathcal{D}^{\text{te}}}} \big[\ell(f_{\theta}, x^*, y^*) - \tau(x^*, y^*) + \tau(x, y) - \ell(f_{\theta}, x, y)\big]  \label{eq:train_test_input_mem_inf_relation}
\end{align}

As in~\citet{sablayrolles2019white}, we make the assumption that $\forall (x,y)\in \mathcal{X}\times\mathcal{Y}$, $\tau(x, y) =\tau$, for some constant value $\tau$. 
 The veracity of this assumption was empirically verified by \citet{sablayrolles2019white}. They showed that membership inference accuracy on CIFAR-10 when using an estimate $\tau(x,y)$ achieves only a 0.5\% improvement over using a constant threshold $\tau$. 
Furthermore, computing $\tau(x,y)$ for each point in the training set is prohibitively expensive for even modestly size datasets, since it requires new models to be trained for each training point.
In theory, the per-sample threshold should outperform a global threshold, however in practice the differences in membership accuracy between the two strategies is small enough as to make virtually no difference. 
To substantiate this claim, we ran membership inference attacks as introduced by \citet{sablayrolles2019white} on MNIST, CIFAR-10 and the Gaussian setting (introduced in \cref{sec:gaussian_membership})  with either the global or local threshold; membership accuracy with the local threshold was always within 0.43\% of accuracy with a global threshold. We believe these small differences imply the assumption is a fair one.

Under this assumption, we reduce \cref{eq:train_test_input_mem_inf_relation} to:

\begin{align}
    \mathop{\mathbb{E}}_{\substack{(x,y)\in\mathcal{D}^{\text{tr}}\\ (x^*,y^*)\in\mathcal{D}^{\text{te}}}} \big[\ell(f_{\theta}, x^*, y^*) - \ell(f_{\theta}, x, y)\big]  \label{eq:simple_train_test_input_mem_inf_relation}
\end{align}

Of course, no information is lost when dividing by a constant and so the membership information leaked by the entire training set in comparison to an equally sized reference test set is found by computing:

\begin{align}
    \mathop{\mathbb{E}}_{\substack{(x,y)\in\mathcal{D}^{\text{tr}}\\ (x^*,y^*)\in\mathcal{D}^{\text{te}}}} \big[ \frac{1}{n}\sum_{i=1}^n\big(\ell(f_{\theta}, x^*_i, y^*_i) - \ell(f_{\theta}, x_i, y_i)\big) \big]  \label{eq:main_train_test_set_mem_inf_relation}
\end{align}

A large absolute value in \cref{eq:main_train_test_set_mem_inf_relation} indicates a severe miscalibration between expected loss on the training and test sets, and so in turn, implies membership information is leaked by the training set. 
Clearly, \cref{eq:main_train_test_set_mem_inf_relation}, is equal to the expected loss gap defined in \cref{sec:introduction}, and is exactly the value we should measure to infer if a model has overfitted on $\mathcal{D}^{\text{tr}}$. 

We have used the terminology set out by~\citet{sablayrolles2019white} to delineate the connection between overfitting and membership inference. 
Contingent on our assumption of equivalent normalizing constants, using
\cref{eq:train_test_input_mem_inf_relation} as a measure for private information leakage directly falls out of the definition of membership inference (c.f. Theorem 1 \& 2 of \citet{sablayrolles2019white}); however other methods such as entropy could also be used in place. We leave this as an interesting direction for future work.

In the following section, we show in a simple setting that adversarial training can provably increase or decrease the risk of overfitting in comparison to standard training. We show how this deficit or excess risk is governed by the size of perturbation used in adversarial training, and the size of training set.

\section{Adversarial training can provably increase or decrease overfitting}
\label{sec:gaussian_membership}

In this section, we identify settings where adversarial training min has a provably larger loss gap in comparison to standard training, for any finitely sized training set. Similarly, we identify settings where adversarial training has a provably smaller loss gap in comparison to standard training, for any finitely sized training set. 
Previous work on the connection between adversarial training and privacy has shown that, empirically, adversarial training increases the efficacy of membership attacks and so, robustness comes at the expense of privacy. In this section, we prove that there exists settings where this trade-off does not exist; one can train a robust model through adversarial training and simultaneously enjoy more privacy in comparison to a standard model.

Let $x\in \mathbb{R}^d$, where $\forall j\in[d]$, $x_j\sim\mathcal{N}(y\mu, \sigma^2)$, $\mu>0$, $\sigma>0$, and $y\sim\{\pm 1\}$ uniformly at random. We study this data model setting under the linear loss function $\ell(f_{\theta}(x), y) = -y\inp{\theta}{x}$, where the decision rule is given by $F_{\theta}(x) = \sign(f_{\theta}(x)) = \sign(\inp{\theta}{x})$. 
We note that robustness properties of a linear classifier optimized with a linear loss or similar variants under Gaussian data have recently been studied by~\citep{chen2020more, min2020curious, yin2018rademacher, tsipras2018robustness, schmidt2018adversarially, nakkiran2019adversarial, khim2018adversarial}~\footnote{Most relevant to our derivations on the differences between the loss gap on standard and robust models is the work of~\citet{chen2020more}. In the same data setting, they show there exists cases where adversarial training can cause the generalization error gap between an adversarially trained robust classifier and a standard classifier to increase and subsequently decrease with more training data, and cases where more training data increases this gap. In~\cref{sec:chen_gen_bounds}, we give an tighter bound for the necessary training set size required to see an increase in this generalization error gap between standard and robust models.}.

In standard and robust empirical risk minimization we compute the following:

\begin{align}
    &\theta_n^{\text{std}} = \argmin_{\theta} \frac{1}{n}\sum^n_{i=1}\ell(f_{\theta}(x_i), y_i) \label{eq: erm} \\
    &\theta_n^{\text{rob}} = \argmin_{\theta} \frac{1}{n}\sum^n_{i=1}\max_{\norm{\delta_i}\leq \epsilon}\ell(f_{\theta}(x_i + \delta_i), y_i) \label{eq: adv_erm}
\end{align}

Under the assumption that the supremum norm of the learned parameters is bounded by $\gamma>0$, \cite{chen2020more} showed that for a linear loss, the exact form of parameters found from standard and robust empirical risk minimization are given by:

\begin{align}
\begin{split}
    \theta_n^{\text{std}} &= \argmin_{\norm{\theta}_{\infty} \leq \gamma} \frac{1}{n}\sum^n_{i=1}-y_i\inp{\theta}{x_i} 
    = \gamma\sign(\sum^n_{i=1}y_ix_i) 
\end{split}
\end{align}
\begin{align}
    \theta_n^{\text{rob}} &= \gamma\sign\big(\sum^n_{i=1}y_ix_i - \epsilon\sign(\sum^n_{i=1}y_ix_i)\big)
\end{align}

Our assumption on the bounded influence of a single training example necessitates that our loss is robust to outliers. Although the linear loss does not exhibit this property for the Gaussian model with large training set sizes, when the number of training points is on the order of $10K-60K$ -- the typical sizes of training sets in practical experiments considered in this work -- the probability that an unusually large $x_i$ determines alone the sign of the corresponding model entry is $<0.1\%$, and this holds for all reasonable choices of $\mu, \sigma$.

To measure how much $\theta_n^{\text{std}}$ and $\theta_n^{\text{rob}}$ overfit, we consider the loss gap as defined in \cref{eq:main_train_test_set_mem_inf_relation}. 
We denote the loss gap between the training set and test set as $r(n)$, and is given by: 

\begin{align}
    r(n) &= \mathop{\mathbb{E}}_{\substack{(x,y)\in\mathcal{D}^{\text{tr}}\\(x^*,y^*)\in\mathcal{D}^{\text{te}}}} \big[ \frac{1}{n}\sum_{i=1}^n\big(\ell(f_{\theta}, x^*_i, y^*_i) - \ell(f_{\theta}, x_i, y_i)\big) \big] \label{eq:standard_gaussian_membership_risk}\\
    &= \frac{1}{n}\mathop{\mathbb{E}}_{\substack{(x,y)\in\mathcal{D}^{\text{tr}}\\(x^*,y^*)\in\mathcal{D}^{\text{te}}}} \big[
    \inp{\theta}{\sum_{i=1}^n x_iy_i} - \inp{\theta}{\sum_{i=1}^n x^*_iy^*_i} \big]
\end{align}

We now proceed to analyze $r(n)$ for both $\theta_n^{\text{std}}$ and $\theta_n^{\text{rob}}$, which we refer to as $r^{\text{std}}(n)$ and $r^{\text{rob}}(n)$, respectively~\footnote{For the interested reader, we provide an interactive plot of $r^{\text{std}}(n)$ and $r^{\text{rob}}(n)$ in \url{https://www.desmos.com/calculator/igkjpul0hz}.}. Full proofs of all claims are given in~\cref{sec:gaussian_membership_proofs}.

\begin{prop}
\label{prop:loss_gap_defined}
The loss gaps, $r^{\text{std}}(n)$ and $r^{\text{rob}}(n)$, are given by:
\begin{align}
    r^{\text{std}}(n) &= d\gamma\sigma\sqrt{\frac{2}{n\pi}}e^{-\frac{n\mu^2}{2\sigma^2}}  \\
    r^{\text{rob}}(n) &= d\gamma\sigma\sqrt{\frac{2}{n\pi}}\bigg(e^{\frac{-n(\epsilon + \mu)^2}{2\sigma^2}} + e^{\frac{-n(\epsilon - \mu)^2}{2\sigma^2}} - e^{\frac{-n\mu^2}{2\sigma^2}}\bigg) 
\end{align}
\end{prop}

\begin{prop}
\label{prop:standard_gaussian_decreasing_prop}
$r^{\text{std}}(n)$ is strictly decreasing in $n$.
\end{prop}

Thus, \cref{prop:standard_gaussian_decreasing_prop} implies that the loss gap under standard training is guaranteed to decrease as the size of training set increases. 
Next, we show this also holds for adversarially robust training.

\begin{prop}
\label{prop:robust_gaussian_decreasing_prop}
For any $\epsilon > 0$, $\lim_{n\rightarrow \infty} r^{\text{rob}}(n) = 0$.
\end{prop}

Thus in the infinite data limit, the loss gap reduces to zero and it is not possible to infer membership of a training input.
From hereon in, we refer to $r^{\text{rob}}(n)$ parameterized by $\epsilon$, as $r^{\text{rob}}(n, \epsilon)$, and analyze $r^{\text{rob}}(n, \epsilon)$ for finite training set sizes, $n$.

\begin{prop}
\label{prop:robust_gaussian_decreasing_roots}
The following hold:
\begin{enumerate}
    \item For $0\leq\epsilon\leq 2\mu$, there is exists no choice of $n\in\mathbb{N}_+$ such that $r^{\text{rob}}(n, \epsilon) = 0$.
    \item For $\epsilon> 2\mu$, $r^{\text{rob}}(n, \epsilon)$ has exactly one real root, $n_0$, that lies in the open set $\bigg( \frac{2\sigma^2\log2}{\epsilon(\epsilon+2\mu)}, \frac{2\sigma^2\log2}{\epsilon(\epsilon-2\mu)} \bigg)$, and there exists $n_1 > n_0$, such that $r^{\text{rob}}(n_1, \epsilon)$ is a minimum.
\end{enumerate}
\end{prop}

\begin{figure}[t]
\centering
\begin{subfigure}{0.199\linewidth}
  \centering
  \includegraphics[width=\linewidth]{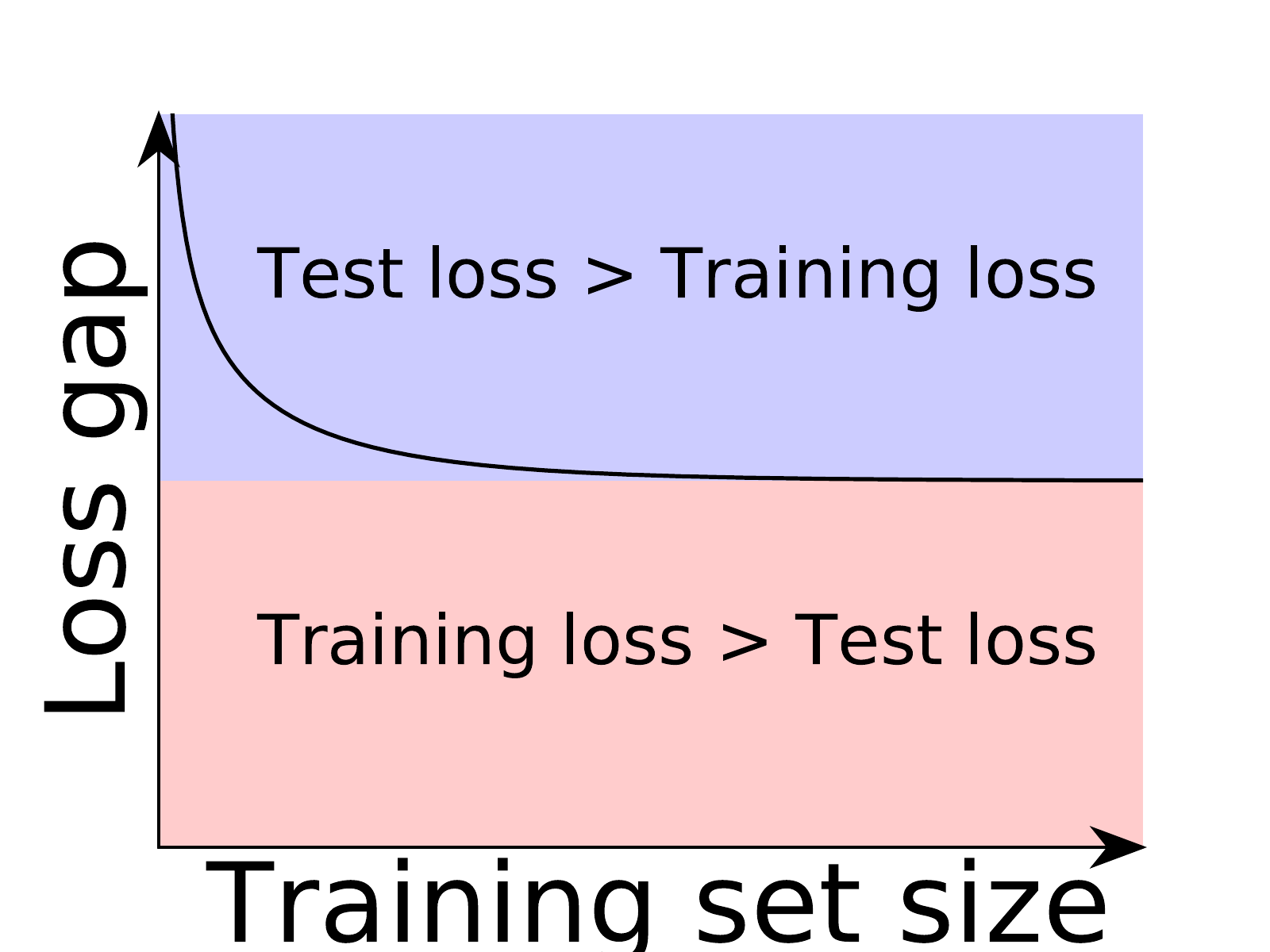}
  \caption{$\epsilon=0$}
  \label{fig:theoretical_gaussian_100d_standard}
\end{subfigure}%
\begin{subfigure}{0.199\linewidth}
  \centering
  \includegraphics[width=\linewidth]{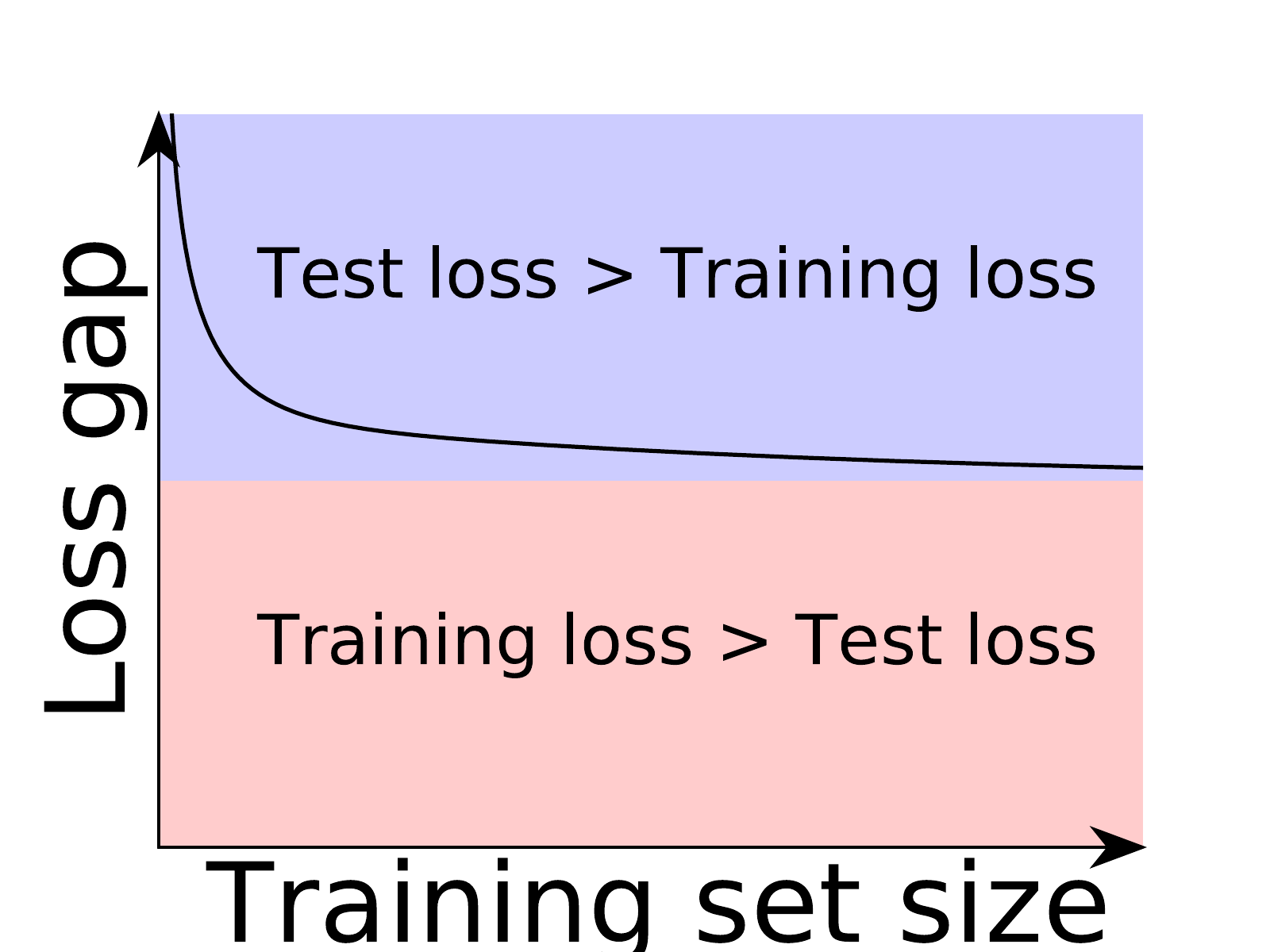}
  \caption{$\epsilon=\frac{\mu}{2}$}
  \label{fig:theoretical_gaussian_100d_weak_adv}
\end{subfigure}%
\begin{subfigure}{0.199\linewidth}
  \centering
  \includegraphics[width=\linewidth]{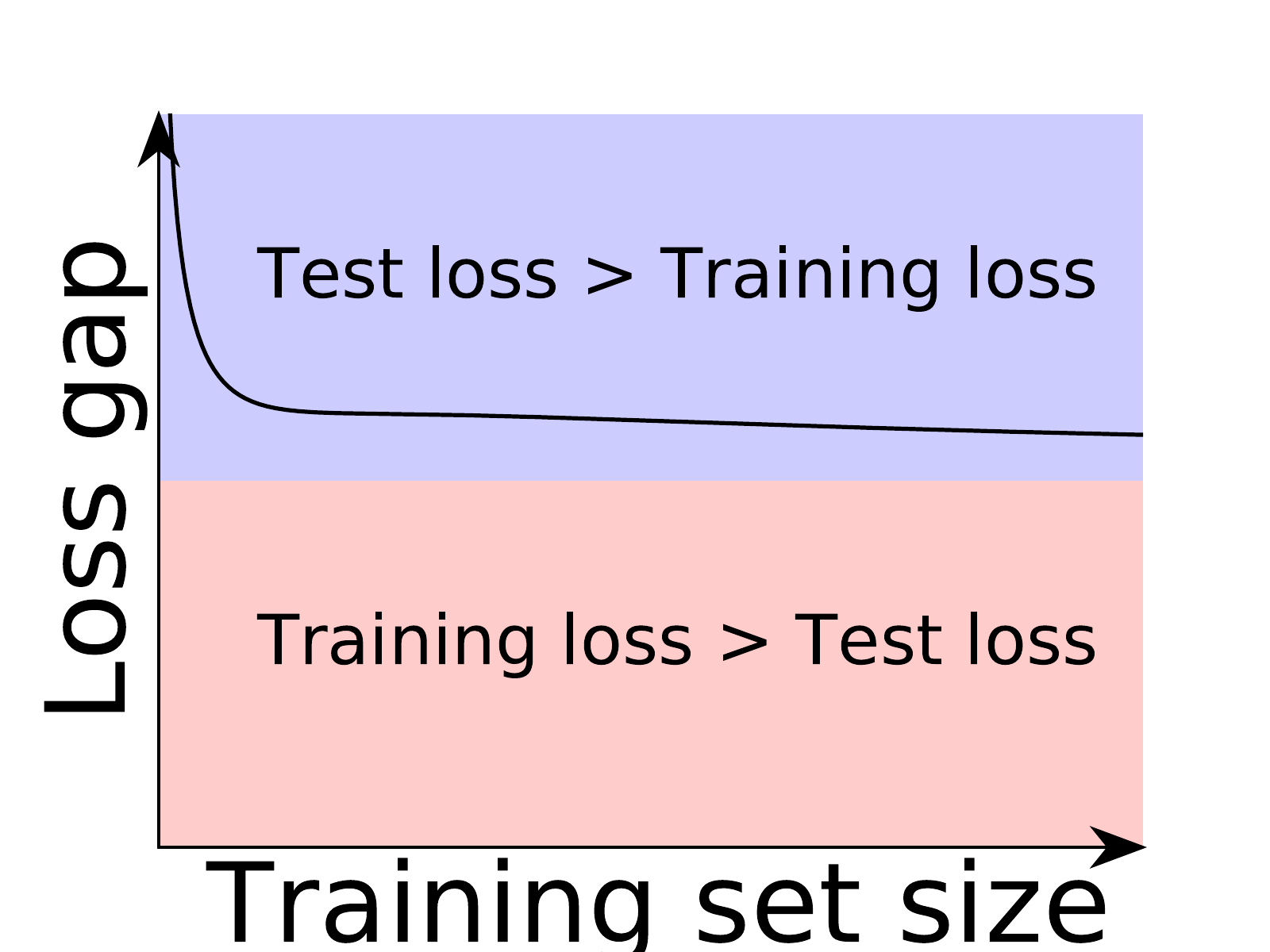}
  \caption{$\epsilon=\mu$}
  \label{fig:theoretical_gaussian_100d_medium_adv}
\end{subfigure}%
\begin{subfigure}{0.199\linewidth}
  \centering
  \includegraphics[width=\linewidth]{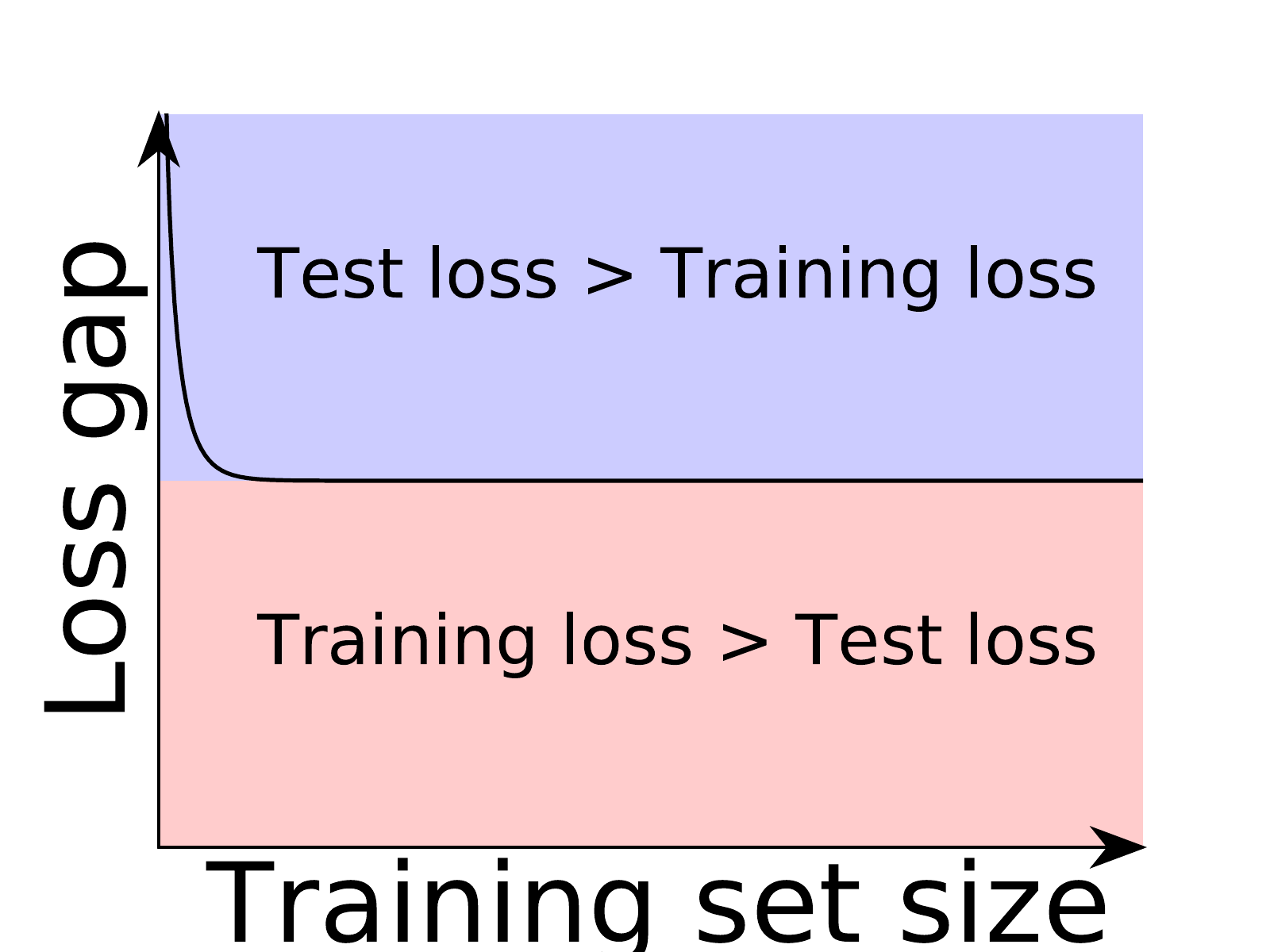}
  \caption{$\epsilon=2\mu$}
  \label{fig:theoretical_gaussian_100d_strong_adv}
\end{subfigure}%
\begin{subfigure}{0.199\linewidth}
  \centering
  \includegraphics[width=\linewidth]{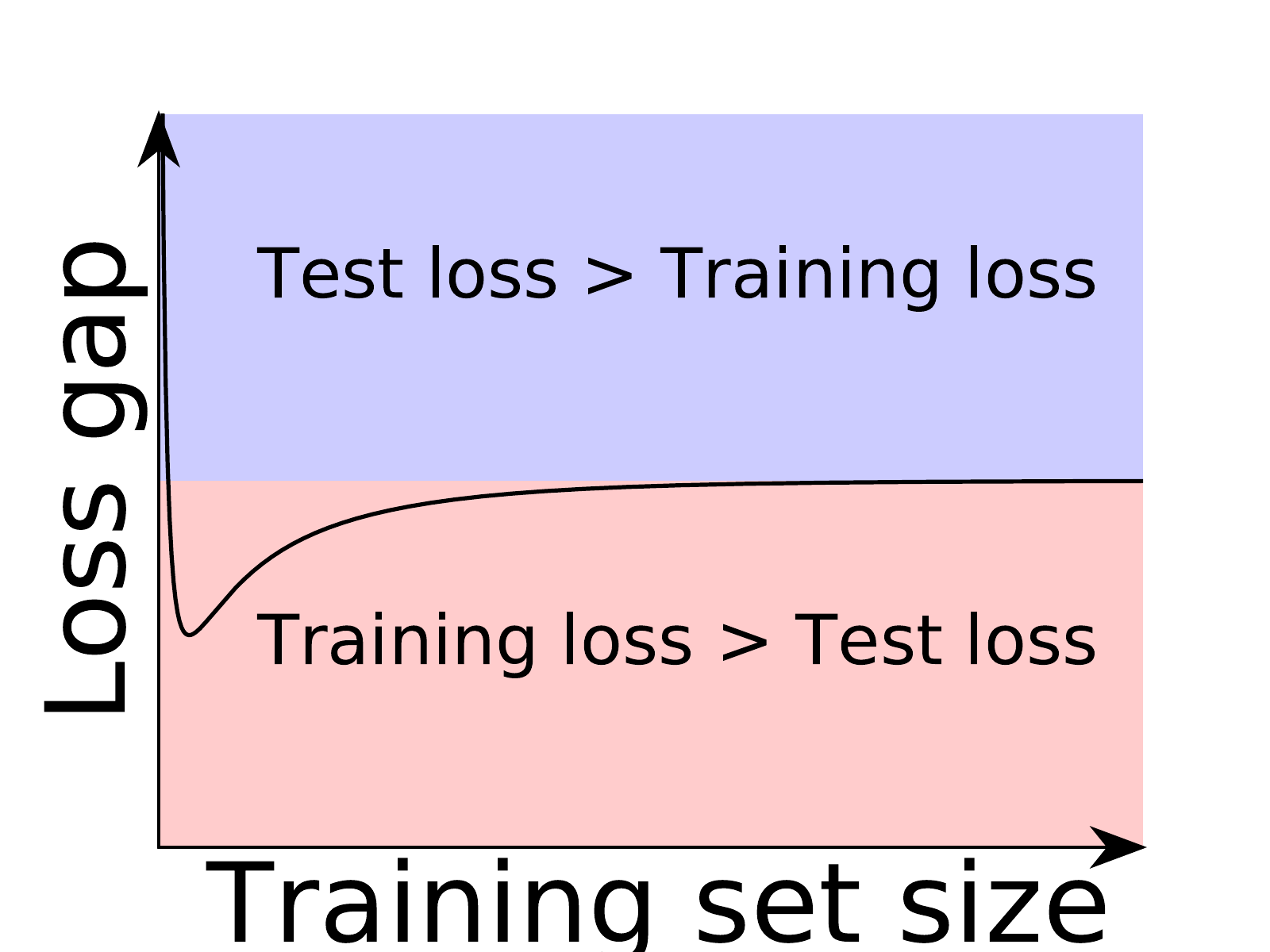}
  \caption{$\epsilon=4\mu$}
  \label{fig:theoretical_gaussian_100d_vstrong_adv}
\end{subfigure}%
\caption{How the loss gap, $r(n)$, decreases as a function of the size of the training set, for different $\epsilon$ used in adversarial training, for $100d$ Gaussian data with $\mu=\sigma=1$. }
\label{fig:theoretical_gaussian_100d}
\end{figure}

\Cref{prop:robust_gaussian_decreasing_roots} implies the if $\epsilon \leq 2\mu$, the loss gap is positive for any finitely sized training set, and so the danger of membership inference is never fully nullified. 
While if $\epsilon > 2\mu$, there exists some $n_0$, such that for any training set larger than $n_0$, the loss gap is negative. 
Of course, this doesn't imply we have a private model; any expected difference in loss between a training and test set implies information has leaked, which can be exploited by an attacker.
Next, we show there exists choices of $n$ and $\epsilon$, such that $r^{\text{rob}}(n, \epsilon)$ is strictly decreasing in $\epsilon$ for training sets smaller than $n$, and strictly increasing for larger training sets.

\begin{prop}
\label{prop:gaussian_r_r0b_dec_or_inc}
For $0<\epsilon<\mu$, $r^{\text{rob}}(n, \epsilon)$ is decreasing in $\epsilon$ for $n<\frac{\sigma^2}{2\mu\epsilon}\log(\frac{\mu+\epsilon}{\mu-\epsilon})$ and increasing in $\epsilon$ for $n>\frac{\sigma^2}{2\mu\epsilon}\log(\frac{\mu+\epsilon}{\mu-\epsilon})$. 
Furthermore, for $\mu<\epsilon<2\mu$, $r^{\text{rob}}(n, \epsilon)$ is decreasing in $\epsilon$ for any training set size, $n$.
\end{prop}

Finally, we now show that there exists cases where $r_n^{\text{rob}} >r_n^{\text{std}}$, and cases where $r_n^{\text{rob}} < r_n^{\text{std}}$.

\begin{prop}
\label{prop:mem_inf_dec_or_inc_with_more_data}
The following hold:
\begin{enumerate}
    \item For $0<\epsilon<\mu$. If $n>\frac{\sigma^2}{2\mu\epsilon}\log(\frac{\mu+\epsilon}{\mu-\epsilon})$, then $r_n^{\text{rob}} >r_n^{\text{std}}$.
    \item If $\epsilon=2\mu$, then $\forall n\in\mathbb{R}_{+}$, $r_n^{\text{rob}} < r_n^{\text{std}}$.
\end{enumerate}
\end{prop}

Thus, if $0<\epsilon<\mu$, robust models overfit more than standard models, and so are more vulnerable to membership attacks, and if $\epsilon=2\mu$, standard models overfit more than robust models, and thus robust models are less vulnerable to membership attacks. 
We visualize this phenomenon in~\cref{fig:theoretical_gaussian_100d}, for a $100d$ Gaussian, with $\mu = \sigma = 1$.
As expected, robust training with $\epsilon=\frac{\mu}{2}$ and $\epsilon=\mu$ (\cref{fig:theoretical_gaussian_100d_weak_adv} and \cref{fig:theoretical_gaussian_100d_medium_adv}, respectively), leads to a larger loss gap in comparison to standard training (\cref{fig:theoretical_gaussian_100d_standard}), while $\epsilon=2\mu$ and $\epsilon=4\mu$ (\cref{fig:theoretical_gaussian_100d_strong_adv} and \cref{fig:theoretical_gaussian_100d_vstrong_adv}, respectively), leads to a smaller loss gap in comparison to standard training. 

\begin{figure}[t]
  \centering
  \includegraphics[width=.7\linewidth]{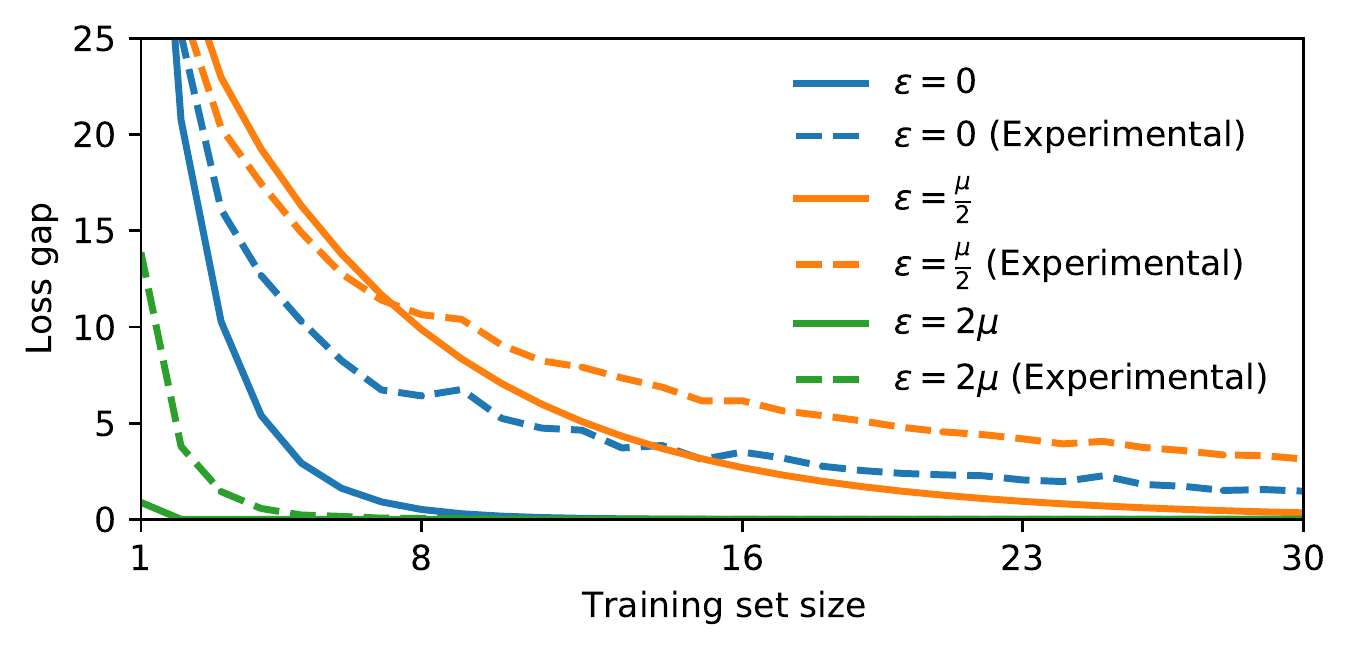}
  \caption{How the loss gap, $r(n)$, decreases as a function of the size of the training set, for different $\epsilon$ used in adversarial training, for $100d$ Gaussian data with $\mu=\sigma=1$. We show both the theoretical curve found from evaluating the closed form solution given in \cref{prop:loss_gap_defined} and the experimental curve found through gradient descent.}
  \label{fig:empirical_gaussian_100d}
\end{figure}

We empirically verified these findings by training a linear classifier with a linear loss on this binary Gaussian problem (with $\mu, \sigma, d$ as defined above) for 200 epochs with gradient descent and a learning rate of 0.001. 
This process was repeated 10$\times$ and we present the averaged results in \cref{fig:empirical_gaussian_100d}. 
The average empirical findings closely match the expected decrease in loss gaps in both standard and robust training.
Of course, a private model that does not achieve a small generalization error is of no practical use; we also measured the average test accuracy of standard training and robust training (for $\epsilon=\frac{\mu}{2}$ and $\epsilon=2\mu$) as the size of training set increases. 
For both standard and robust training, the average generalization error fell to zero if $n>2$. 
Thus, it is possible to achieve zero generalization error while incurring a non-zero loss gap, we explore why in~\cref{sec:bayes_risk_analysis}. 
This loss gap decreases as the size of the training set increases, and consequently, privacy of the model increases if there is more available training data.
Furthermore, it is not necessarily the case that robust models are \emph{less} private, we have shown there exists $\epsilon$ and $n$ where robust models have a smaller loss gap and thus are \emph{more} private than standard models.

\section{Experimental results}
\label{sec:cifar10_results}

In~\cref{sec:gaussian_membership} we described the relationship between overfitting and robust training and how it is dependent on the size of the training set.
However this formal connection was only proved on a simple binary classification problem. 
Here, we experimentally validate that these results hold for more complex classification problems. 
We also empirically demonstrate the strong correlation between the loss gap and membership attack accuracy, as suggested by \cref{eq:main_train_test_set_mem_inf_relation}.

To expose the relationship between the loss gap and training set size, we train a ResNet-18 classifier~\citep{he2016deep} on the CIFAR-10 and CIFAR-100 dataset~\citep{krizhevsky2009learning}, for various training set sizes. 
We then empirically show that an increase in overfitting increases membership attack accuracy. 
We compare standard training with adversarially robust training with FGSM~\citep{goodfellow2014explaining} with $\epsilon=\nicefrac{8}{255}$, and adversarially robust training with PGD~\citep{madry2017towards} with $\epsilon=\nicefrac{8}{255}$ and ten attack iterations. 
In all experiments, the initial learning rate was set to 0.1 and was annealed to 0.01 and 0.001, at epochs 100 and 150, respectively. 
During training, we apply data augmentation by randomly cropping and flipping inputs.

\begin{figure*}[t]
\centering
\begin{subfigure}{.35\linewidth}
  \centering
  \includegraphics[width=\linewidth]{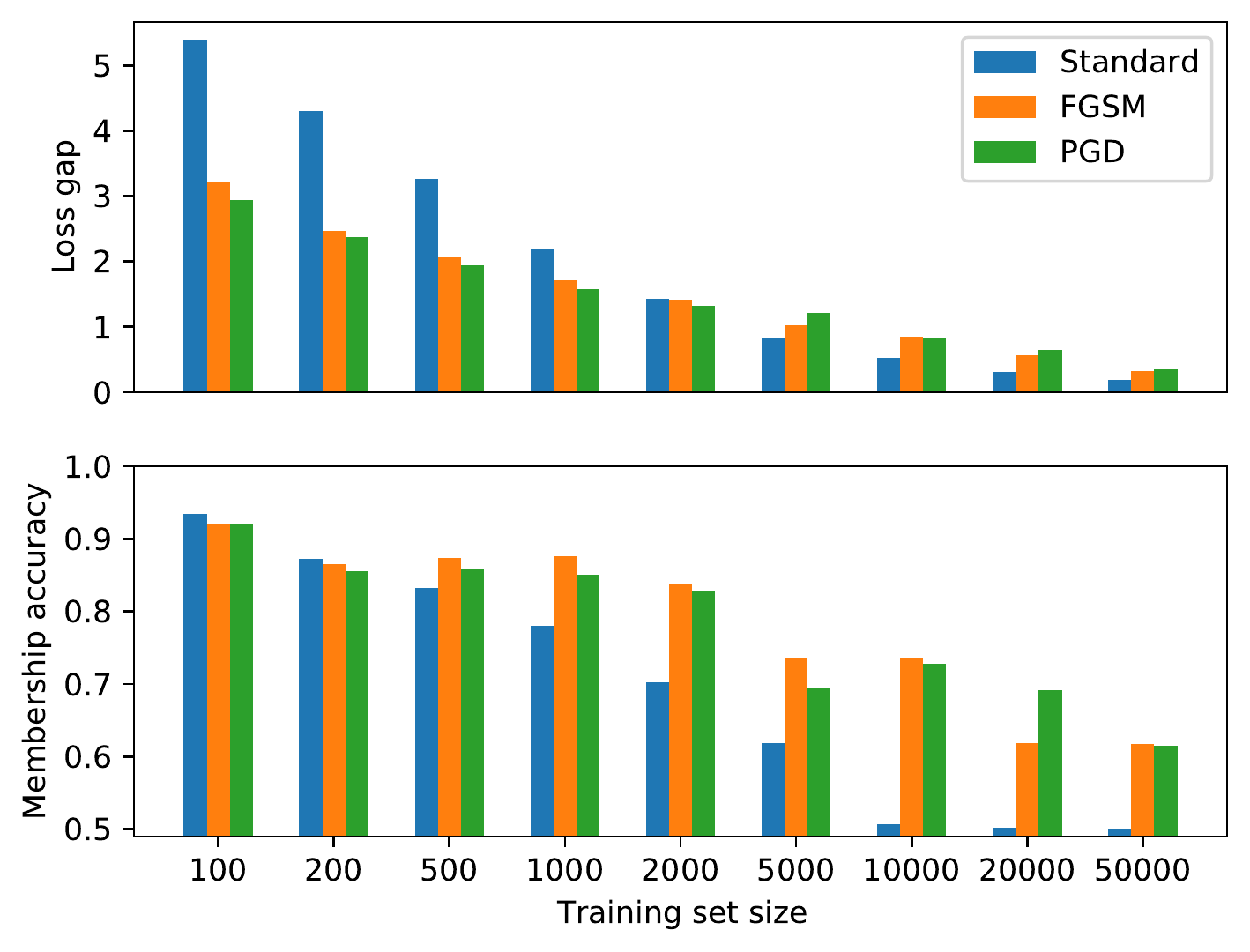}
  \caption{CIFAR-10.}
  \label{fig:cifar10_mem_inf_train_vs_size_median_transforms}
\end{subfigure}%
\begin{subfigure}{.35\linewidth}
  \centering
  \includegraphics[width=\linewidth]{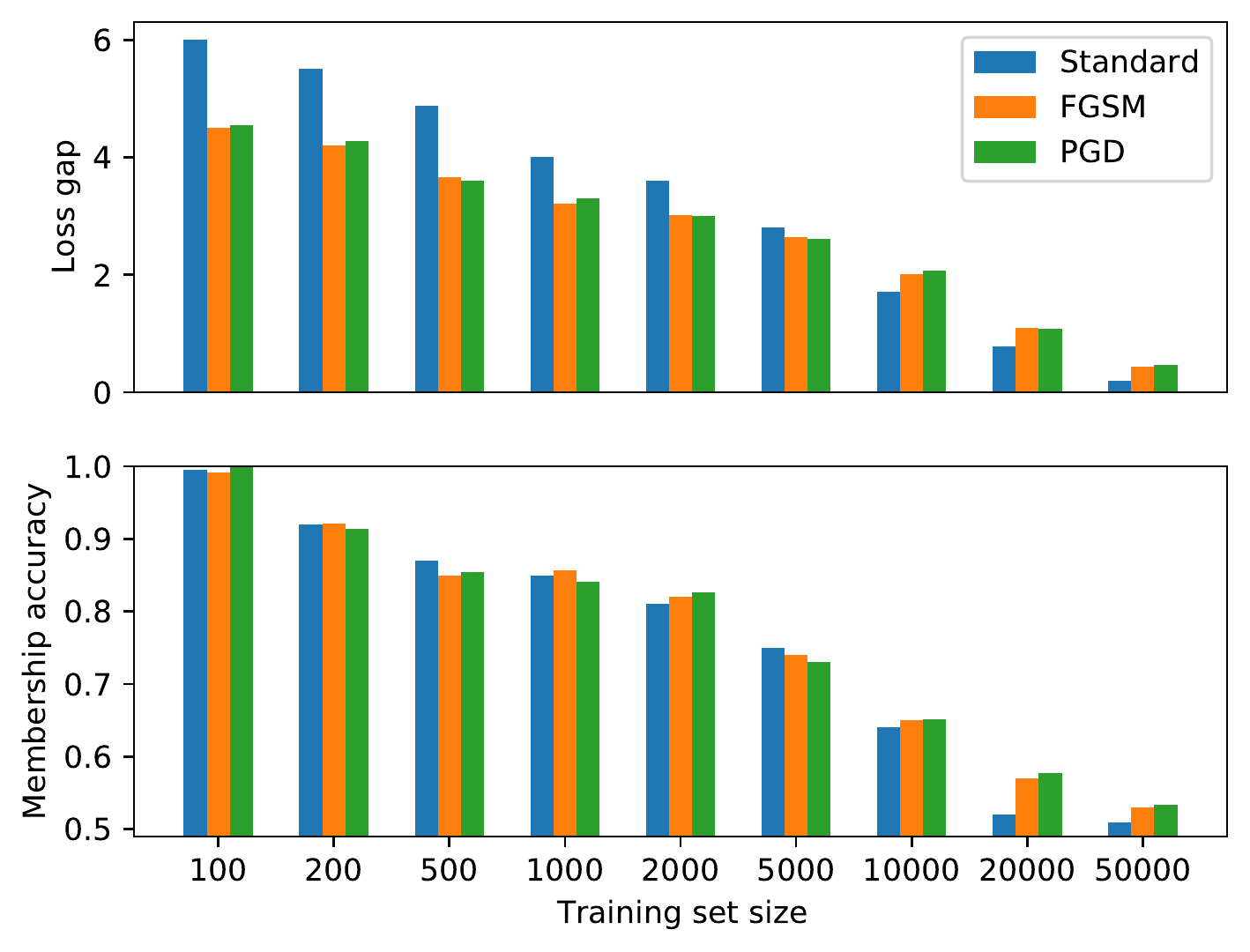}
  \caption{CIFAR-100.}
  \label{fig:cifar100_mem_inf_train_vs_size_median_transforms}
\end{subfigure}
\caption{Membership attack accuracy and loss gap as a function of the training set size, for standard and robust models (using adversarial training with $\epsilon=\frac{8}{255}$ and using either the FGSM or PGD attack). We use the MALT membership attack~\citep{sablayrolles2019white} with a median decision threshold as explained in~\cref{sec:cifar10_results}.}
\label{fig:cifar10_mem_inf_train_vs_size}
\end{figure*}

We use a simplified version of the MALT attack as described in~\citet{sablayrolles2019white} for the membership attack.
To perform the attack, the adversary first estimates the constant loss threshold, $\tau$, described in \cref{sec:membership_inference_and_overfitting}, as follows: 
The adversary trains their own model on a dataset that is different (but from a similar distribution) from data on which the attack is mounted. 
The adversary then finds the \emph{median} loss for both the training and test sets on which they trained and evaluated their own model, and then finds the loss that maximizes the average distance between these two values. 
This is then used as threshold to decide if an input was or was not part of the training set in the attack.

As an aside,~\citet{sablayrolles2019white} uses a \emph{mean} loss threshold, however  we found that using the mean threshold in a MALT attack performs poorly in comparison to a median threshold, if large outliers occur when recording the threshold.
For example, in standard training with a training and test set size of 100, the median threshold is 2.7 while the mean threshold is 0.01. 
The training loss values are heavily concentrated around zero, while the test loss values have mean 3.7 but varied between 0 and 19.1. 
Thus a median threshold performed better than a mean threshold since this gave no false negatives. 
Although the mean threshold attack is the default choice among related work, it performs poorly in comparison to the median threshold, since there is a saturation of data points with a loss just above zero. We report attack results using the mean loss threshold in \cref{fig:cifar10_mem_inf_train_vs_size_mean_threshold} of \cref{sec:cifar10_results_appendix}.

\citet{sablayrolles2019white} showed that this simple threshold attack improves upon other membership attacks~\citep{shokri2017membership, yeom2018privacy}. 
We measure the membership accuracy on training and test sets of equal size; because the CIFAR-10 and CIFAR-100 data has 10K test set inputs, when we train with larger training set sizes, we randomly select a subset of the training set that is equal in size to the test set to measure membership accuracy.
We repeat this process 5$\times$, and report the average accuracy. 
For experiments with smaller training and test sets, we also repeat the experiment 5$\times$ selecting random subsets from the full training and test sets and average the results.

\Cref{fig:cifar10_mem_inf_train_vs_size} shows the average membership attack accuracy and the corresponding loss gaps for different training set sizes. As expected from our analysis in~\cref{sec:gaussian_membership}, the training set size is tightly correlated with both membership accuracy and the loss gap -- as the training set size increases both of these values decreases. 
Furthermore, we expect from our analysis in \cref{sec:gaussian_membership}, that for a small $\epsilon$ (we use  $\epsilon=\nicefrac{8}{255}$) in robust training, the loss gap (and thus membership accuracy) decreases at a slower rate in comparison to standard training. 
This property is exactly what is exhibited in~\cref{fig:cifar10_mem_inf_train_vs_size}, for small training set sizes, the loss gap under standard training is larger or equivalent to the loss gap in robust training, and consequently the membership accuracy is also equivalent. 
However as the training set size increases, the loss gap and membership accuracy on standard models falls more rapidly than on robust models.
For completeness, we plot the membership accuracy, the loss gap, training accuracy, and test accuracy throughout training for each training set size in~\cref{sec:cifar10_results_appendix}.
Note, that there do exist cases in our experiments with larger loss gaps but smaller membership accuracy, for example \cref{fig:cifar10_mem_inf_train_vs_size_median_transforms} with training set size 500. We believe this is mainly due to the fact that membership attacks are currently quite weak, and so cannot fully reveal the correspondence between the loss gap and membership vulnerability -- see \citet{choo2020label}, where they show membership inference attacks exploiting the loss gap are only 10\% better than a naive attack that decides membership based on if the classifier predicts an incorrect class. However,the experiments on CIFAR-100 in \cref{fig:cifar100_mem_inf_train_vs_size_median_transforms} show a stronger correspondence between loss and membership in this dataset, where a large/small loss gap leads to large/small membership vulnerability.
\section{Conclusion}
\label{sec:conclusion}

Ostensibly, a weakness of this work is that the robust optimization procedure creates examples that should be classified differently by an oracle classifier, and so the term ``adversarial examples'' is a misnomer. 
We fully agree that this analysis blurs the meaning of the term ``adversarial examples'', however we do not make any claims surrounding ground-truth classification of adversarial examples. 
Indeed, we are entirely uninterested in adversarial examples in this work; what we are concerned with is how robust optimization procedures (such as adversarial training) affect other desirable properties of a classifier, such as privacy (resistance to membership inference attacks).
Of course, we relate privacy to the $\epsilon$ used during the robust optimization procedure, but we are not interested in the robust generalization error, as was the main focus in \citet{tsipras2018there} and 
\citet{tsipras2018robustness} -- we are interested solely in two properties, standard error and privacy. 
Interestingly, in the Gaussian model, standard error fell to zero for even very large $\epsilon$ values, and so it is fair to measure these models with equivalent test set accuracy, in terms of their privacy. 

We show in \cref{fig:cifar10_mem_inf_train_vs_size} that $\epsilon$ does not need to be large to train models that are more private than standard models. For example, using a CIFAR-10 training set of size between 200-500, and setting $\epsilon=\frac{8}{255}$ -- which is small enough that perturbed inputs will maintain their ground-truth semantic label -- the robust models are more resistant to privacy attacks than standard models. Thus, in practice, we do not need to use a large $\epsilon$ to inherit this privacy property in robust models. Translating our theory from a Gaussian setting to more interesting distributions to study this property in more detail is an interesting direction for future work. 

In sum, this work formally described the trade-offs between private and robust machine learning. 
In a simple data setting, we proved that a robust model can either be more \emph{or} less private than a standard model, and so there exists settings where there is no trade-off at all. 
We showed that this trade-off, or lack thereof, depends entirely on the size of the training set and the strength of the adversarial perturbations used during the training of a robust model.
We then showed our findings can be observed on more complex datasets, such as CIFAR-10 and CIFAR-100.

\bibliography{main}

\clearpage
\onecolumn
\appendix
\section{Deferred proofs: Adversarial training can provably increase or decrease overfitting}
\label{sec:gaussian_membership_proofs}

Here, we give full proofs for claims made in~\cref{sec:gaussian_membership}.

\begin{repprop}{prop:loss_gap_defined}
The loss gap, $r^{\text{std}}(n)$ and $r^{\text{rob}}(n)$, are given by:
\begin{align}
    r^{\text{std}}(n) &= d\gamma\sigma\sqrt{\frac{2}{n\pi}}e^{-\frac{n\mu^2}{2\sigma^2}}  \\
    r^{\text{rob}}(n) &= d\gamma\sigma\sqrt{\frac{2}{n\pi}}\bigg(e^{\frac{-n(\epsilon + \mu)^2}{2\sigma^2}} + e^{\frac{-n(\epsilon - \mu)^2}{2\sigma^2}} - e^{\frac{-n\mu^2}{2\sigma^2}}\bigg) 
\end{align}
\end{repprop}

\begin{proof}

Let $u\coloneqq \frac{1}{n}\sum_{i=1}^n x_iy_i$, then if $\theta \coloneqq \theta_n^{\text{std}}$, we have

\begin{align}
    r^{\text{std}}(n) &= \frac{1}{n}\mathop{\mathbb{E}}_{\substack{(x,y)\in\mathcal{D}^{\text{tr}}\\(x^*,y^*)\in\mathcal{D}^{\text{te}}}} \big[
    \inp{\gamma\sign(u)}{\sum_{i=1}^n x_iy_i} - \inp{\gamma\sign(u)}{\sum_{i=1}^n x^*_iy^*_i} \big] \\
    &= \frac{1}{n}\sum_{j=1}^d\mathop{\mathbb{E}}_{\substack{(x_{j},y)\in\mathcal{D}^{\text{tr}}\\(x_{j}^*,y^*)\in\mathcal{D}^{\text{te}}}} \big[
     \inp{\gamma\sign(u_j)}{\sum_{i=1}^n x_{ij}y_{i}} - \inp{\gamma\sign(u_j)}{\sum_{i=1}^n x^*_{ij}y^*_{i}} \big] \label{eq:standard_risk_d_dim}
\end{align}

Since, given an input, $x_i$, each dimension $x_{ij}$, is drawn from $\mathcal{N}(y_i\mu, \sigma^2)$, \cref{eq:standard_risk_d_dim} is equal to $d$
copies over a single dimension. In the following derivations we omit the dimensional subscript, but stress all expectations are now taken over a single dimension:

\begin{align}
    r^{\text{std}}(n) &= \frac{d}{n}\mathop{\mathbb{E}}_{\substack{(x,y)\in\mathcal{D}^{\text{tr}}\\(x^*,y^*)\in\mathcal{D}^{\text{te}}}} \big[
    \inp{\gamma\sign(u)}{\sum_{i=1}^n x_iy_i} - \inp{\gamma\sign(u)}{\sum_{i=1}^n x^*_iy^*_i} \big] \\
    &= d\gamma \mathop{\mathbb{E}}_{u\sim\mathcal{N}(\mu, \frac{\sigma^2}{n})}\big[ u\sign(u)\big] -
    \frac{d\gamma}{n} \mathop{\mathbb{E}}_{u\sim\mathcal{N}(\mu, \frac{\sigma^2}{n})}\big[\sign(u)\big]\cdot\mathop{\mathbb{E}}_{\substack{ (x_i^*,y_i^*)\in\mathcal{D}^{\text{te}}}}\big[\sum_{i=1}^n x^*_iy^*_i\big]  \label{eq:standard_risk_complicated}
\end{align}

To solve \cref{eq:standard_risk_complicated}, we note that $\mathbb{E}[u\sign(u)] = \mathbb{E}[\abs{u}]$, and:

\begin{align}
    &\mathop{\mathbb{E}}_{u\sim\mathcal{N}(\mu, \frac{\sigma^2}{n})}[\sign(u)] = P(u>0)-P(u<0) 
    = 2\Phi(\frac{\sqrt{n}\mu}{\sigma}) - 1 \\
    &\mathop{\mathbb{E}}_{u\sim\mathcal{N}(\mu, \frac{\sigma^2}{n})}[\abs{u}] = \sigma\sqrt{\frac{2}{n\pi}}e^{-\frac{n\mu^2}{2\sigma^2}}-\mu\big(1-2\Phi(\frac{\sqrt{n}\mu}{\sigma})\big) \\
    &\mathop{\mathbb{E}}_{(x^*,y^*)\in\mathcal{D}^{\text{te}}}\big[\sum_{i=1}^n x^*_iy^*_i\big] = n\mu
\end{align}

Thus for standard empirical risk minimization, $\theta_n^{\text{std}}$, the overfitting measure defined in~\cref{eq:main_train_test_set_mem_inf_relation} is equal to:

\begin{align}
    r^{\text{std}}(n) &= d\gamma\big(\sigma\sqrt{\frac{2}{n\pi}}e^{-\frac{n\mu^2}{2\sigma^2}}-\mu\big(1-2\Phi(\frac{\sqrt{n}\mu}{\sigma})\big)\big) -\frac{d\gamma}{n}\big(n\mu\big(2\Phi(\frac{\sqrt{n}\mu}{\sigma}) - 1\big)\big) \\
    &= d\gamma\sigma\sqrt{\frac{2}{n\pi}}e^{-\frac{n\mu^2}{2\sigma^2}}
\end{align}

Similarly, for robust empirical risk minimization, $\theta_n^{\text{rob}}$, the overfitting measure defined in~\cref{eq:main_train_test_set_mem_inf_relation} is equal to:

\begin{align}
    r^{\text{rob}}(n) = d\gamma\mathop{\mathbb{E}}_{u\sim\mathcal{N}(\mu, \frac{\sigma^2}{n})}[u\sign(u-\epsilon\sign(u))] -d\gamma\mu\mathop{\mathbb{E}}_{u\sim\mathcal{N}(\mu, \frac{\sigma^2}{n})}[\sign(u-\epsilon\sign(u))] \label{eq:mem_inf_rob_gaussian}
\end{align}

To find an analytical form of \cref{eq:mem_inf_rob_gaussian}, we must find closed form solutions to $\mathbb{E}[\sign(u-\epsilon\sign(u))]$ and $\mathbb{E}[u\sign(u-\epsilon\sign(u))]$:

\begin{align}
    \mathop{\mathbb{E}}_{u\sim\mathcal{N}(\mu, \frac{\sigma^2}{n})}[\sign(u-\epsilon\sign(u))] &= P(u<\epsilon\sign(u)) - P(u>\epsilon\sign(u)) \\
    &= 1 + 2P(u<0) -2P(u<\epsilon) -2P(u<-\epsilon) \\
    &= 1 + 2\Phi(-\frac{\sqrt{n}\mu}{\sigma}) - 2\Phi(\frac{\sqrt{n}(\epsilon-\mu)}{\sigma}) - 2\Phi(\frac{\sqrt{n}(-\epsilon-\mu)}{\sigma}) \\
    \mathop{\mathbb{E}}_{u\sim\mathcal{N}(\mu, \frac{\sigma^2}{n})}[u\sign(u-\epsilon\sign(u))] 
    &= \int_{-\infty}^{-\epsilon} \frac{-u\sqrt{n}}{\sqrt{2\pi}\sigma}e^{\frac{-n(u-\mu)^2}{2\sigma^2}}du + \int_{-\epsilon}^{0} \frac{u\sqrt{n}}{\sqrt{2\pi}\sigma}e^{\frac{-n(u-\mu)^2}{2\sigma^2}}du \\
    &+ \int_{0}^{\epsilon} \frac{-u\sqrt{n}}{\sqrt{2\pi}\sigma}e^{\frac{-n(u-\mu)^2}{2\sigma^2}}du + \int_{\epsilon}^{\infty} \frac{u\sqrt{n}}{\sqrt{2\pi}\sigma}e^{\frac{-n(u-\mu)^2}{2\sigma^2}}du \\
    &= \sigma\frac{2}{n\pi}\big[e^{\frac{-n(\epsilon + \mu)^2}{2\sigma^2}} + e^{\frac{-n(\epsilon - \mu)^2}{2\sigma^2}} - e^{\frac{-n\mu^2}{2\sigma^2}}\big] \\
    &-\mu\big[ -1 - 2\Phi(-\frac{\sqrt{n}\mu}{\sigma}) + 2\Phi(\frac{\sqrt{n}(\epsilon-\mu)}{\sigma}) + 2\Phi(\frac{\sqrt{n}(-\epsilon-\mu)}{\sigma}) \big]
\end{align}

Thus the loss gap as defined in~\cref{eq:mem_inf_rob_gaussian} is equal to:

\begin{align}
    r^{\text{rob}}(n) &= d\gamma\sigma\sqrt{\frac{2}{n\pi}}\bigg(e^{\frac{-n(\epsilon + \mu)^2}{2\sigma^2}} + e^{\frac{-n(\epsilon - \mu)^2}{2\sigma^2}} - e^{\frac{-n\mu^2}{2\sigma^2}}\bigg) \\
    &-d\gamma\mu\bigg( -1 - 2\Phi(-\frac{\sqrt{n}\mu}{\sigma}) + 2\Phi(\frac{\sqrt{n}(\epsilon-\mu)}{\sigma}) + 2\Phi(\frac{\sqrt{n}(-\epsilon-\mu)}{\sigma})\bigg) \\
    &-d\gamma\mu\bigg( 1 + 2\Phi(-\frac{\sqrt{n}\mu}{\sigma}) - 2\Phi(\frac{\sqrt{n}(\epsilon-\mu)}{\sigma}) - 2\Phi(\frac{\sqrt{n}(-\epsilon-\mu)}{\sigma})\bigg) \\
    &= d\gamma\sigma\sqrt{\frac{2}{n\pi}}\bigg(e^{\frac{-n(\epsilon + \mu)^2}{2\sigma^2}} + e^{\frac{-n(\epsilon - \mu)^2}{2\sigma^2}} - e^{\frac{-n\mu^2}{2\sigma^2}}\bigg)
\end{align}

\end{proof}

\begin{repprop}{prop:standard_gaussian_decreasing_prop}
$r^{\text{std}}(n)$ is strictly decreasing in $n$.
\end{repprop}

\begin{proof}
Let $r^{\text{std}}(n) = d\gamma\sigma\sqrt{\frac{2}{n\pi}} e^{\frac{-n\mu^2}{2\sigma^2}}$. Then, 

\begin{align}
    \frac{\partial r^{\text{std}}}{\partial n} = \frac{-d\gamma\mu^2}{\sigma\sqrt{2n\pi}}e^{\frac{-n\mu^2}{2\sigma^2}} - \frac{d\gamma\sigma}{\sqrt{2\pi}n^{\frac{3}{2}}}e^{\frac{-n\mu^2}{2\sigma^2}} 
\end{align}

Since $d, n,\mu,\sigma,\gamma > 0$, then $\frac{\partial r^{\text{std}}}{\partial n} < 0$, $\forall n\in(0,\infty)$, and so $r^{\text{std}}(n)$ is decreasing on $\mathbb{R}_{+}$. 
It is also the case that $\mathop{\lim}_{n\rightarrow 0^{+}}r^{\text{std}}(n)=\infty$ as $\mathop{\lim}_{n\rightarrow 0^{+}}e^{\frac{-n\mu^2}{2\sigma^2}}= 1$ and $\mathop{\lim}_{n\rightarrow 0^{+}}\frac{1}{\sqrt{n}}=\infty$. 
Similarly, $\mathop{\lim}_{n\rightarrow \infty}r^{\text{std}}(n)=0$ as $\mathop{\lim}_{n\rightarrow \infty}e^{\frac{-n\mu^2}{2\sigma^2}}=0$.
\end{proof}

\begin{repprop}{prop:robust_gaussian_decreasing_prop}
For any $\epsilon > 0$, $\lim_{n\rightarrow \infty} r^{\text{rob}}(n) = 0$.
\end{repprop}

\begin{proof}
$r^{\text{rob}}(n) =  d\gamma\sigma\sqrt{\frac{2}{n\pi}}\bigg(e^{\frac{-n(\epsilon + \mu)^2}{2\sigma^2}} + e^{\frac{-n(\epsilon - \mu)^2}{2\sigma^2}} - e^{\frac{-n\mu^2}{2\sigma^2}}\bigg)$, then

\begin{align}
    \frac{\partial r^{\text{rob}}}{\partial n} = \frac{d}{\sqrt{2\pi}\sigma n^{\frac{3}{2}}}\bigg(&n\gamma\big(-(\epsilon + \mu)^2e^{\frac{-n(\epsilon+\mu)^2}{2\sigma^2}} - (\epsilon - \mu)^2e^{\frac{-n(\epsilon-\mu)^2}{2\sigma^2}} +\mu^2e^{\frac{-n\mu^2}{2\sigma^2}}\big) \\
    -&\gamma\sigma^2\big(
    e^{\frac{-n(\epsilon+\mu)^2}{2\sigma^2}} + e^{\frac{-n(\epsilon-\mu)^2}{2\sigma^2}} -e^{\frac{-n\mu^2}{2\sigma^2}}\big)\bigg)
\end{align}

Since $d,\mu,\sigma,\gamma,\epsilon > 0$, we have $\lim_{n\rightarrow \infty} r^{\text{rob}}(n)=0$, since $\lim_{n\rightarrow \infty} e^{\frac{-n(\epsilon+\mu)^2}{2\sigma^2}} = \lim_{n\rightarrow \infty} e^{\frac{-n(\epsilon-\mu)^2}{2\sigma^2}} = \lim_{n\rightarrow \infty} e^{\frac{-n\mu^2}{2\sigma^2}} = 0$.
\end{proof}

\begin{repprop}{prop:robust_gaussian_decreasing_roots}
The following hold:
\begin{enumerate}
    \item For $0\leq\epsilon\leq 2\mu$, there is exists no choice of $n\in\mathbb{N}_+$ such that $r^{\text{rob}}(n, \epsilon) = 0$.
    \item For $\epsilon> 2\mu$, $r^{\text{rob}}(n, \epsilon)$ has exactly one real root, $n_0$, that lies in the open set $\bigg( \frac{2\sigma^2\log2}{\epsilon(\epsilon+2\mu)}, \frac{2\sigma^2\log2}{\epsilon(\epsilon-2\mu)} \bigg)$, and there exists $n_1 > n_0$, such that $r^{\text{rob}}(n_1, \epsilon)$ is a minimum.
\end{enumerate}
\end{repprop}

\begin{proof}

We have that, for a fixed $\epsilon>0$, $r^{\text{rob}}(n, \epsilon)=0$ either when $n=\infty$ or when $e^{\frac{-n(\epsilon + \mu)^2}{2\sigma^2}} + e^{\frac{-n(\epsilon - \mu)^2}{2\sigma^2}} - e^{\frac{-n\mu^2}{2\sigma^2}} = 0$. Thus, there exists a real root of $r^{\text{rob}}(n, \epsilon) \iff e^{\frac{-n}{2\sigma^2}\epsilon(\epsilon + 2\mu)} + e^{\frac{-n}{2\sigma^2}\epsilon(\epsilon - 2\mu)} = 1$. Let $x\coloneqq e^{\frac{n}{2\sigma^2}}$.
Thus, a real root exists if the following holds:

\begin{align}
    \label{eq: simple_g_zero_condition}
    x^{-\epsilon(\epsilon + 2\mu)} + x^{-\epsilon(\epsilon - 2\mu)} = 1
\end{align}

Let $h(x, \epsilon) = x^{-\epsilon(\epsilon + 2\mu)} + x^{-\epsilon(\epsilon - 2\mu)} - 1$. We have the following limit equalities:

\begin{align}
    &\lim_{x\rightarrow\infty} x^{-\epsilon(\epsilon + 2\mu)} = 0\\
    &\lim_{x\rightarrow\infty} x^{-\epsilon(\epsilon - 2\mu)} = 
    \begin{cases}
    \infty, & \text{if } 0<\epsilon<2\mu \\
    1, & \text{if } \epsilon=2\mu \\
    0, & \text{if } \epsilon>2\mu \\
    \end{cases}\\
    &\lim_{x\rightarrow\infty} h(x, \epsilon) = 
    \begin{cases}
    \infty, & \text{if } 0<\epsilon<2\mu \\
    0, & \text{if } \epsilon=2\mu \\
    -1, & \text{if } \epsilon>2\mu \\
  \end{cases}
\end{align}

We also note for any $\epsilon>0$, $\lim_{x\rightarrow 0^+} h(x, \epsilon) = \infty$.
The derivative of $h(x, \epsilon)$ is given by:

\begin{align}
    \frac{\partial h}{\partial x} = -\epsilon(\epsilon + 2\mu)x^{-\epsilon(\epsilon + 2\mu)-1} -\epsilon(\epsilon - 2\mu)x^{-\epsilon(\epsilon - 2\mu)-1}
\end{align}

$h(x,\epsilon)$ decreasing is equivalent to $\frac{\partial h}{\partial x} < 0$. In turn this implies:

\begin{align}
    \frac{\partial h}{\partial x} < 0 &\implies -\epsilon(\epsilon + 2\mu)x^{-\epsilon(\epsilon + 2\mu)-1} < \epsilon(\epsilon - 2\mu)x^{-\epsilon(\epsilon - 2\mu)-1} \\
    &\implies -x^{-4\mu\epsilon} < \frac{\epsilon - 2\mu}{\epsilon + 2\mu} \\
    &\implies \bigg(\frac{2\mu + \epsilon}{2\mu-\epsilon}\bigg)^{\frac{1}{4\mu\epsilon}} < x \label{eq:h_zero_cond_1}
\end{align}

\noindent \textit{Proof of (2)}.

\Cref{eq:h_zero_cond_1} holds $\forall\epsilon\in(0,\infty)$ when $2\mu<\epsilon$ since $\frac{2\mu + \epsilon}{2\mu-\epsilon} < 0 $ and $x^{4\mu\epsilon}>0$. 
Thus $h(x,\epsilon)$ is decreasing on $(0, \infty)$ when $2\mu<\epsilon$ and has a real root since $\lim_{x\rightarrow 0^+} h(x,\epsilon) = \infty$ and $\lim_{x\rightarrow \infty} h(x,\epsilon) = -1$. 

We can find lower and upper bounds for this root when $\epsilon>2\mu$.
Note, $h(x,\epsilon)<2x^{-\epsilon(\epsilon -2\mu)} - 1$ when $\epsilon>2\mu$.
So $x = 2^{\frac{1}{\epsilon(\epsilon-2\mu)}}$ gives an upper bound.
Similarly, $h(x,\epsilon)>2x^{-\epsilon(\epsilon +2\mu)} - 1$ when $\epsilon>2\mu$, and $x = 2^{\frac{1}{\epsilon(\epsilon+2\mu)}}$ gives a lower bound.

Since $x=e^{\frac{n}{2\sigma^2}}$, there exists some $n_0>0$, such that this $n_0$ lies in $\bigg( \frac{2\sigma^2\log2}{\epsilon(\epsilon+2\mu)}, \frac{2\sigma^2\log2}{\epsilon(\epsilon-2\mu)} \bigg)$ and is a root of $r^{\text{rob}}(n_0, \epsilon)=0$. Note that $\forall\epsilon>2\mu$, $h(n,\epsilon)$ has a real root and $\lim_{n\rightarrow 0^+}h(n,\epsilon)=\infty$ and $\lim_{n\rightarrow \infty}h(n,\epsilon)=0$.
Thus there exists some $n_1>n_0$ that is a minimum of $h$ and $n_0$ gives a trivial lower bound.

\noindent \textit{Proof of (1)}.

Firstly, if $\epsilon=0$ or $\epsilon=2\mu$, then $r^{\text{rob}}(n, 2\mu) = \sqrt{\frac{2}{n\pi}}\gamma\sigma e^{\frac{-3n\mu^2}{2\sigma^2}}$ or $r^{\text{std}}(n, 0) = \sqrt{\frac{2}{n\pi}}\gamma\sigma e^{\frac{-n\mu^2}{2\sigma^2}}$, respectively. Clearly, for any finite $n$, $r^{\text{rob}}(n, 0) \neq 0$ and $r^{\text{rob}}(n, 2\mu) \neq 0$.

For $0<\epsilon<2\mu$, we can show $h(x,\epsilon)$ has no real root on $(0,\infty)$ if we show $h$ has a global single minimum at $x_0\in(0,\infty)$ and $h(x_0,\epsilon)>0$. The minimum of $h(x,\epsilon)$ is given by:

\begin{align}
    \frac{\partial h}{\partial x} = 0 
    \implies x_0 = \bigg(\frac{2\mu+\epsilon}{2\mu-\epsilon}\bigg)^{\frac{1}{4\mu\epsilon}}
\end{align}

We also have $\lim_{x\rightarrow 0^+} h(x,\epsilon) = \lim_{x\rightarrow {\infty}} h(x,\epsilon) = \infty$. 
First, we show that $h(x_0, \epsilon)$ achieves its maximal value as $\epsilon\rightarrow 0^+$. 
That is, $h(x_0, \epsilon)$ is decreasing on $0<\epsilon<2\mu$.
Following this we show $\lim_{\epsilon\rightarrow 0^+}h(x_0, \epsilon) > 0$,  and $\lim_{\epsilon\rightarrow 2\mu}h(x_0, \epsilon) = 0$. 
This shows there exists no real roots if $0<\epsilon<2\mu$.

To show $h(x_0, \epsilon)$ is decreasing on $0<\epsilon<2\mu$, we show $\frac{\partial h(x_0, \epsilon)}{\partial \epsilon} < 0$. 

Note that $h(x_0,\epsilon) = \bigg(\frac{2\mu-\epsilon}{2\mu+\epsilon}\bigg)^{\frac{\epsilon + 2\mu}{4\mu}} + \bigg(\frac{2\mu-\epsilon}{2\mu+\epsilon}\bigg)^{\frac{\epsilon - 2\mu}{4\mu}} - 1$, and:

\begin{align}
\frac{\partial h(x_0, \epsilon)}{\partial \epsilon} = \big(\frac{a}{b}\big)^{\frac{-a}{4\mu}}\bigg(\frac{ab(\frac{a}{b^2} + \frac{1}{b})}{4\mu a} + \frac{\log(\frac{a}{b})}{4\mu}\bigg) + 
\big(\frac{a}{b}\big)^{\frac{b}{4\mu}}\bigg(\frac{-b^2(\frac{a}{b^2} + \frac{1}{b})}{4\mu a} + \frac{\log(\frac{a}{b})}{4\mu}\bigg)
\end{align}

where $a=2\mu -\epsilon$ and $b=2\mu + \epsilon$. Then,

\begin{align}
    \frac{\partial h(x_0, \epsilon)}{\partial \epsilon} < 0 
    &\implies \frac{a}{b}\big(\frac{a}{b} + \log(\frac{a}{b}) + 1\big) < \frac{b}{a} - \log(\frac{a}{b}) + 1 \\
    &\implies a\log(\frac{a}{b}) + b\log(\frac{a}{b}) < 0 \label{eq:h_zero_cond_2} 
\end{align}

Note \cref{eq:h_zero_cond_2} holds since $0<a<b$ and $\log(\frac{a}{b})<0$.
Hence $h(x_0, \epsilon)$ is decreasing in $\epsilon$ and the maximum value is found at $\lim_{\epsilon\rightarrow 0^+} h(x_0, \epsilon) = \lim_{\epsilon\rightarrow 0^+} \bigg(\frac{2\mu-\epsilon}{2\mu+\epsilon}\bigg)^{\frac{\epsilon + 2\mu}{4\mu}} + \bigg(\frac{2\mu-\epsilon}{2\mu+\epsilon}\bigg)^{\frac{\epsilon - 2\mu}{4\mu}} - 1$. Note:

\begin{align}
    &\lim_{\epsilon\rightarrow 0^+} \bigg(\frac{2\mu-\epsilon}{2\mu+\epsilon}\bigg)^{\frac{\epsilon + 2\mu}{4\mu}} = \lim_{\epsilon\rightarrow 0^+} \bigg(\frac{2\mu-\epsilon}{2\mu+\epsilon}\bigg)^{\frac{\epsilon - 2\mu}{4\mu}} = 1
\end{align}

Thus $\lim_{\epsilon\rightarrow 0^+} h(x_0, \epsilon)=1$, and:

\begin{align}
    &\lim_{\epsilon\rightarrow 2\mu} \bigg(\frac{2\mu-\epsilon}{2\mu+\epsilon}\bigg)^{\frac{\epsilon + 2\mu}{4\mu}} = 0 \\
    &\lim_{\epsilon\rightarrow 2\mu} \bigg(\frac{2\mu-\epsilon}{2\mu+\epsilon}\bigg)^{\frac{\epsilon - 2\mu}{4\mu}} = 1
\end{align}

So $\lim_{\epsilon\rightarrow 2\mu} h(x_0, \epsilon)=0$, and thus $h(x, \epsilon)$ does not have a root on $x\in(0, \infty)$ and $\epsilon\in(0, 2\mu)$.

\end{proof}

\begin{repprop}{prop:gaussian_r_r0b_dec_or_inc}
For $0<\epsilon<2\mu$, $r^{\text{rob}}(n, \epsilon)$ is decreasing in $\epsilon$ for $n<\frac{\sigma^2}{2\mu\epsilon}\log(\frac{\mu+\epsilon}{\mu-\epsilon})$ and increasing in $\epsilon$ for $n>\frac{\sigma^2}{2\mu\epsilon}\log(\frac{\mu+\epsilon}{\mu-\epsilon})$.
\end{repprop}

\begin{proof}

\begin{align}
    \frac{\partial r^{\text{rob}}(n, \epsilon)}{\partial \epsilon}
    = \frac{d\gamma}{\sigma}\sqrt{\frac{2n}{\pi}}\big((\mu-\epsilon)e^{\frac{-n(\mu-\epsilon)^2}{2\sigma^2}} - (\mu+\epsilon)e^{\frac{-n(\mu+\epsilon)^2}{2\sigma^2}}\big)
\end{align}

Firstly we note that if $\mu < \epsilon < 2\mu$, then $\frac{\partial r^{\text{rob}}(n, \epsilon)}{\partial n}<0$ since $\mu-\epsilon < 0$ and $-(\mu+\epsilon) < 0$.
We also note $\frac{\partial r^{\text{rob}}(n, \epsilon_1)}{\partial \epsilon}<\frac{\partial r^{\text{rob}}(n, \epsilon_2)}{\partial \epsilon}$ when $\mu<\epsilon_1<\epsilon_2<2\mu$.
Hence it is decreasing in $\epsilon$ for $\mu<\epsilon<2\mu$ for every $n>0$.

For $0<\epsilon<\mu$, we have $\frac{\partial r^{\text{rob}}(n, \epsilon)}{\partial \epsilon}<0$ when:

\begin{align}
    &(\mu-\epsilon)e^{\frac{-n(\mu-\epsilon)^2}{2\sigma^2}} < (\mu+\epsilon)e^{\frac{-n(\mu+\epsilon)^2}{2\sigma^2}} \\
    \implies &\frac{2n\mu\epsilon}{\sigma^2}<\log(\frac{\mu + \epsilon}{\mu - \epsilon})
\end{align}

and similarly $\frac{\partial r^{\text{rob}}(n, \epsilon)}{\partial \epsilon}>0$ when

\begin{align}
    \frac{2n\mu\epsilon}{\sigma^2}>\log(\frac{\mu + \epsilon}{\mu - \epsilon})
\end{align}

Thus, for $0<\epsilon<\mu$, if $n<\frac{\sigma^2}{2\mu\epsilon}\log(\frac{\mu+\epsilon}{\mu-\epsilon})$, then $r^{\text{rob}}(n, \epsilon)$ decreases as $\epsilon$ increases, and if $n>\frac{\sigma^2}{2\mu\epsilon}\log(\frac{\mu+\epsilon}{\mu-\epsilon})$,  $r^{\text{rob}}(n, \epsilon)$ increases as $\epsilon$ increases.

\end{proof}

\begin{repprop}{prop:mem_inf_dec_or_inc_with_more_data}
The following hold:
\begin{enumerate}
    \item For $0<\epsilon<\mu$. If $n>\frac{\sigma^2}{2\mu\epsilon}\log(\frac{\mu+\epsilon}{\mu-\epsilon})$, then $r_n^{\text{rob}} >r_n^{\text{std}}$.
    \item If $\epsilon=2\mu$, then $\forall n\in\mathbb{R}_{+}$, $r_n^{\text{rob}} < r_n^{\text{std}}$.
\end{enumerate}
\end{repprop}

\begin{proof}
\begin{enumerate}
    \item Firstly, at $\epsilon=0$, $r_n^{\text{rob}} = r_n^{\text{std}}$. From~\cref{prop:gaussian_r_r0b_dec_or_inc}, we also have that if $n>\frac{\sigma^2}{2\mu\epsilon}\log(\frac{\mu+\epsilon}{\mu-\epsilon})$, then $r_n^{\text{rob}}$ is increasing as $\epsilon\rightarrow\mu$. It follows then that for $n>\frac{\sigma^2}{2\mu\epsilon}\log(\frac{\mu+\epsilon}{\mu-\epsilon})$, $r_n^{\text{rob}} >r_n^{\text{std}}$.
    \item For $\epsilon=2\mu$,
    \begin{align}
        r_n^{\text{rob}} &= d\gamma\sigma\sqrt{\frac{2}{n\pi}}\bigg(e^{\frac{-3n\mu^2}{2\sigma^2}} + e^{\frac{-n\mu^2}{2\sigma^2}} - e^{\frac{-n\mu^2}{2\sigma^2}}\bigg) \\
        &= d\gamma\sigma\sqrt{\frac{2}{n\pi}} e^{\frac{-3n\mu^2}{2\sigma^2}} \\
        r_n^{\text{std}} &= d\gamma\sigma\sqrt{\frac{2}{n\pi}} e^{\frac{-n\mu^2}{2\sigma^2}}
    \end{align}
    Now $\forall n\in\mathbb{R}_{+}$, $e^{\frac{-3n\mu^2}{2\sigma^2}} < e^{\frac{-n\mu^2}{2\sigma^2}}$, and so $r_n^{\text{rob}} < r_n^{\text{std}}$.
\end{enumerate}
\end{proof}
\clearpage
\section{More experimental results}
\label{sec:cifar10_results_appendix}

\begin{figure}[!htb]
\centering
\begin{subfigure}{.33\textwidth}
  \centering
  \includegraphics[width=\linewidth]{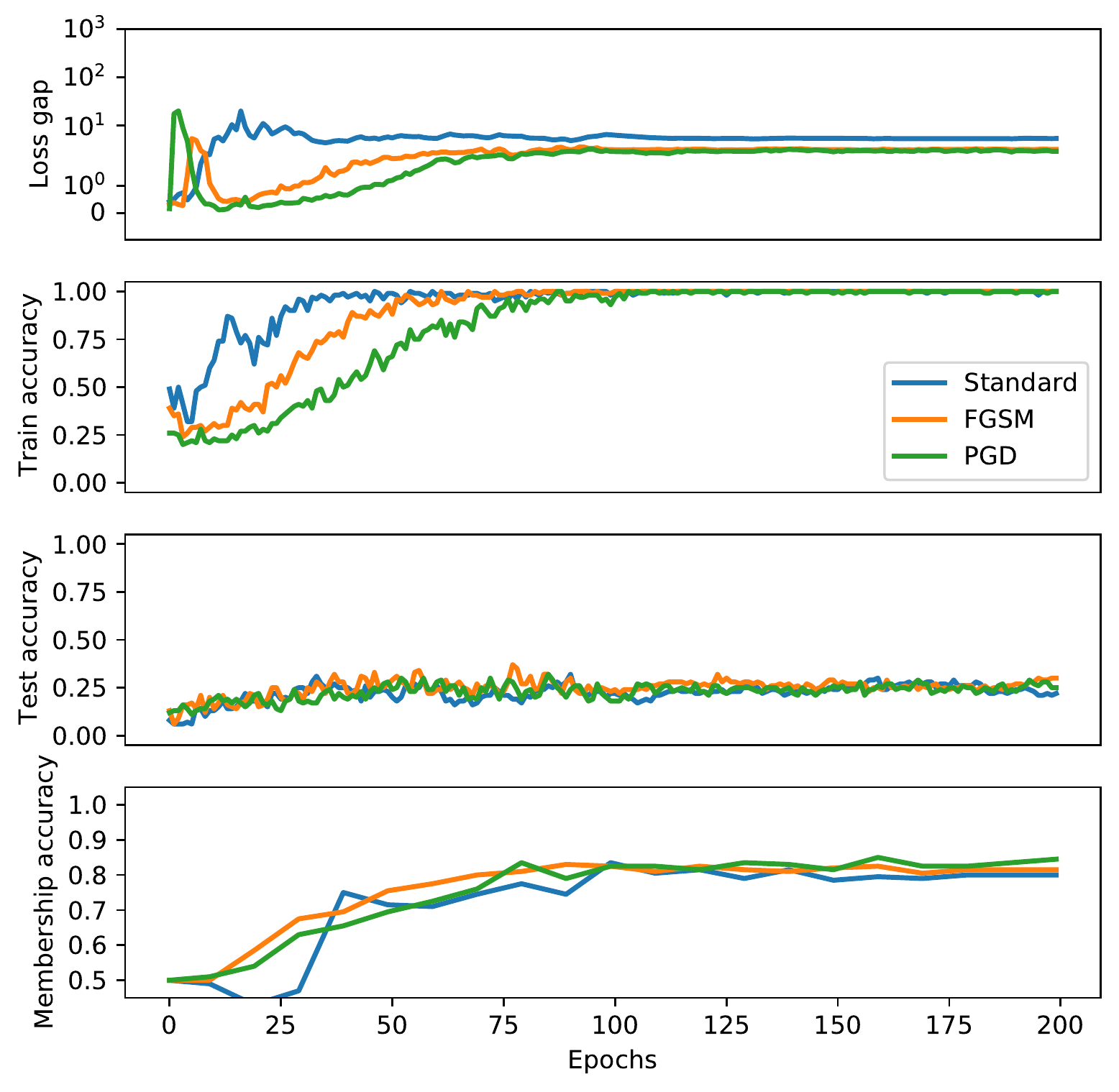}
  \caption{Training set size: 100.}
  \label{fig:cifar10_mem_inf_train_vs_epoch_mean_transform_100}
\end{subfigure}%
\begin{subfigure}{.33\textwidth}
  \centering
  \includegraphics[width=\linewidth]{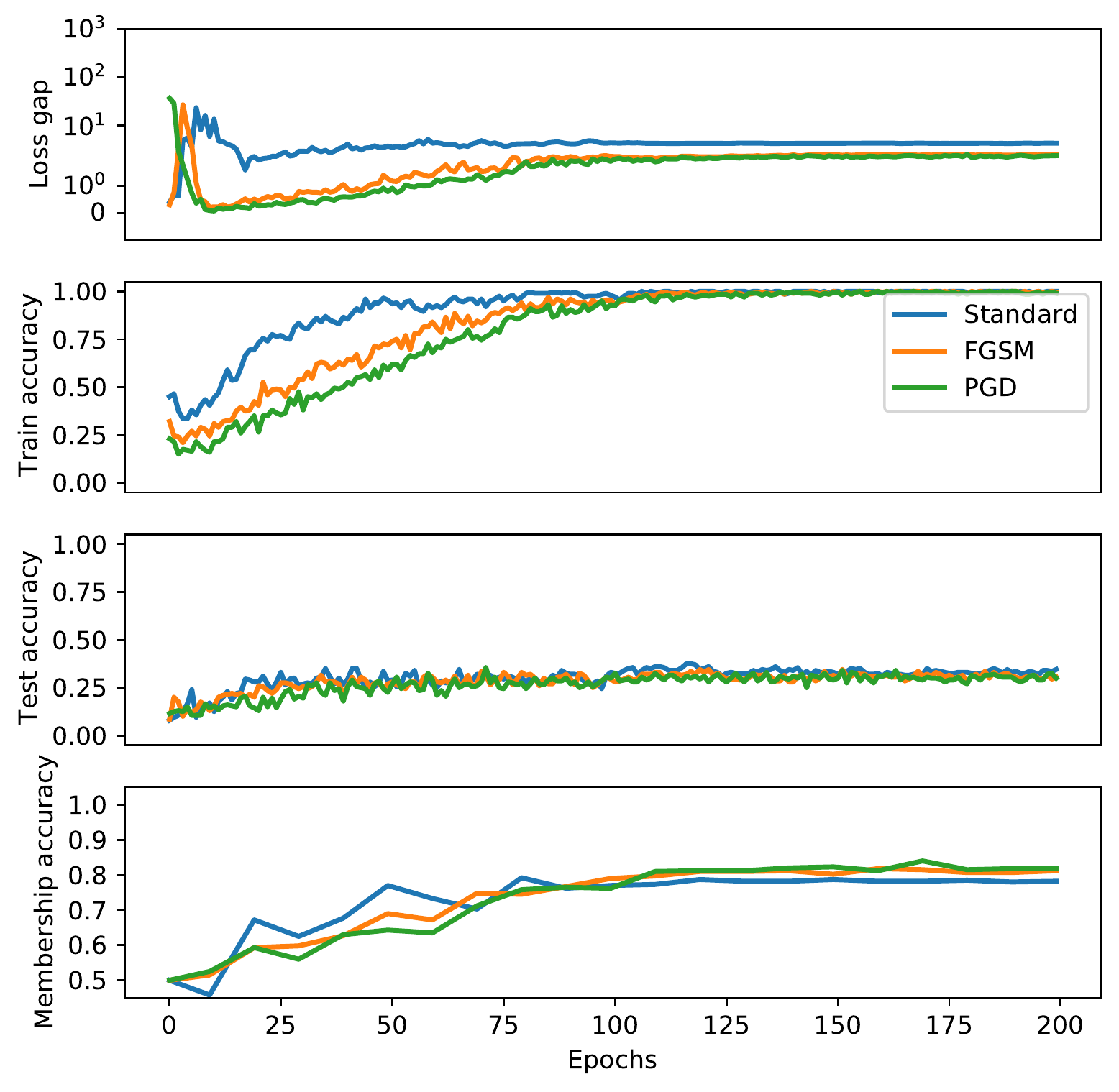}
  \caption{Training set size: 200.}
  \label{fig:cifar10_mem_inf_train_vs_epoch_mean_transform_200}
\end{subfigure}%
\begin{subfigure}{.33\textwidth}
  \centering
  \includegraphics[width=\linewidth]{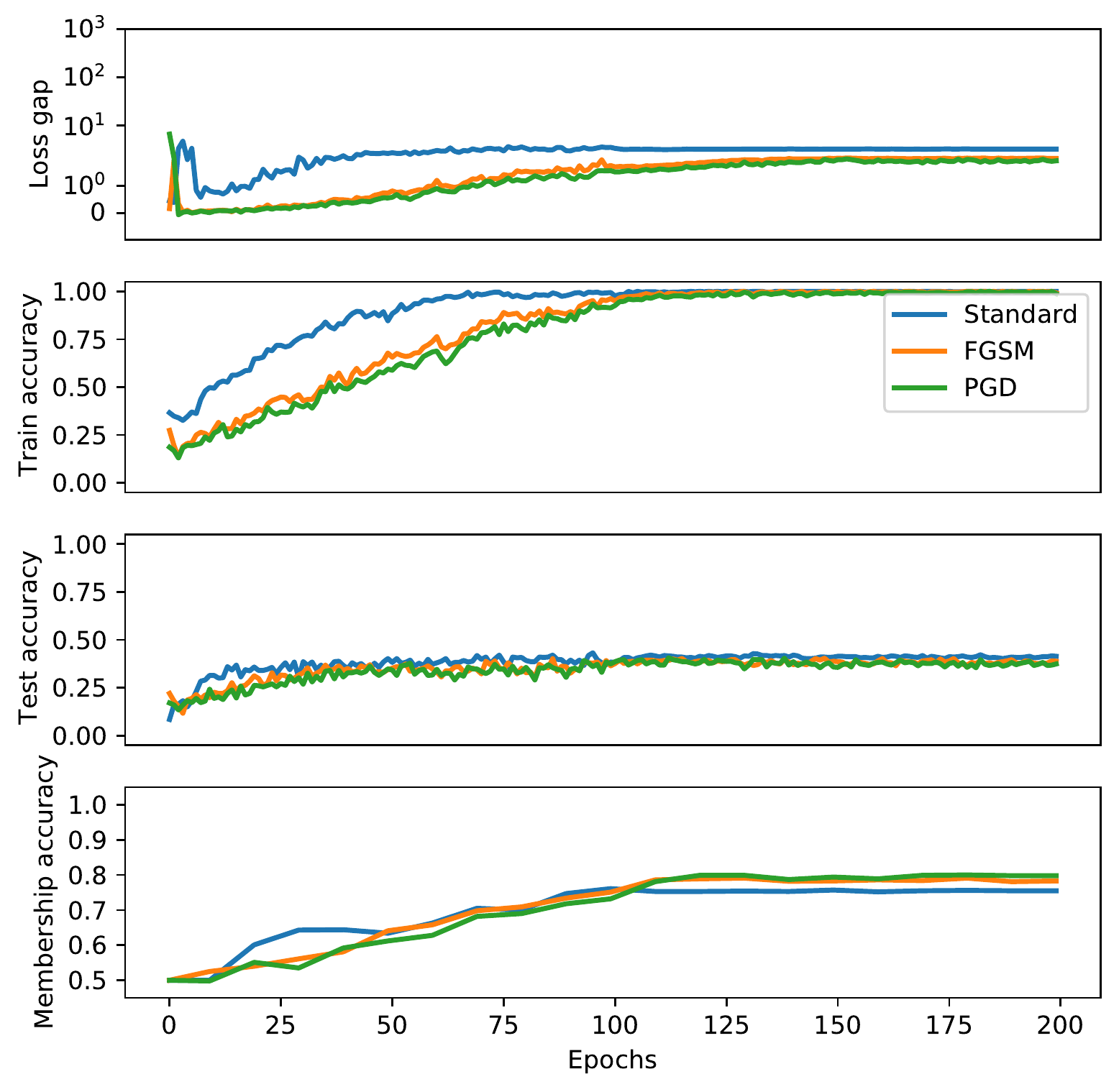}
  \caption{Training set size: 500.}
  \label{fig:cifar10_mem_inf_train_vs_epoch_mean_transform_500}
\end{subfigure}

\begin{subfigure}{.33\textwidth}
  \centering
  \includegraphics[width=\linewidth]{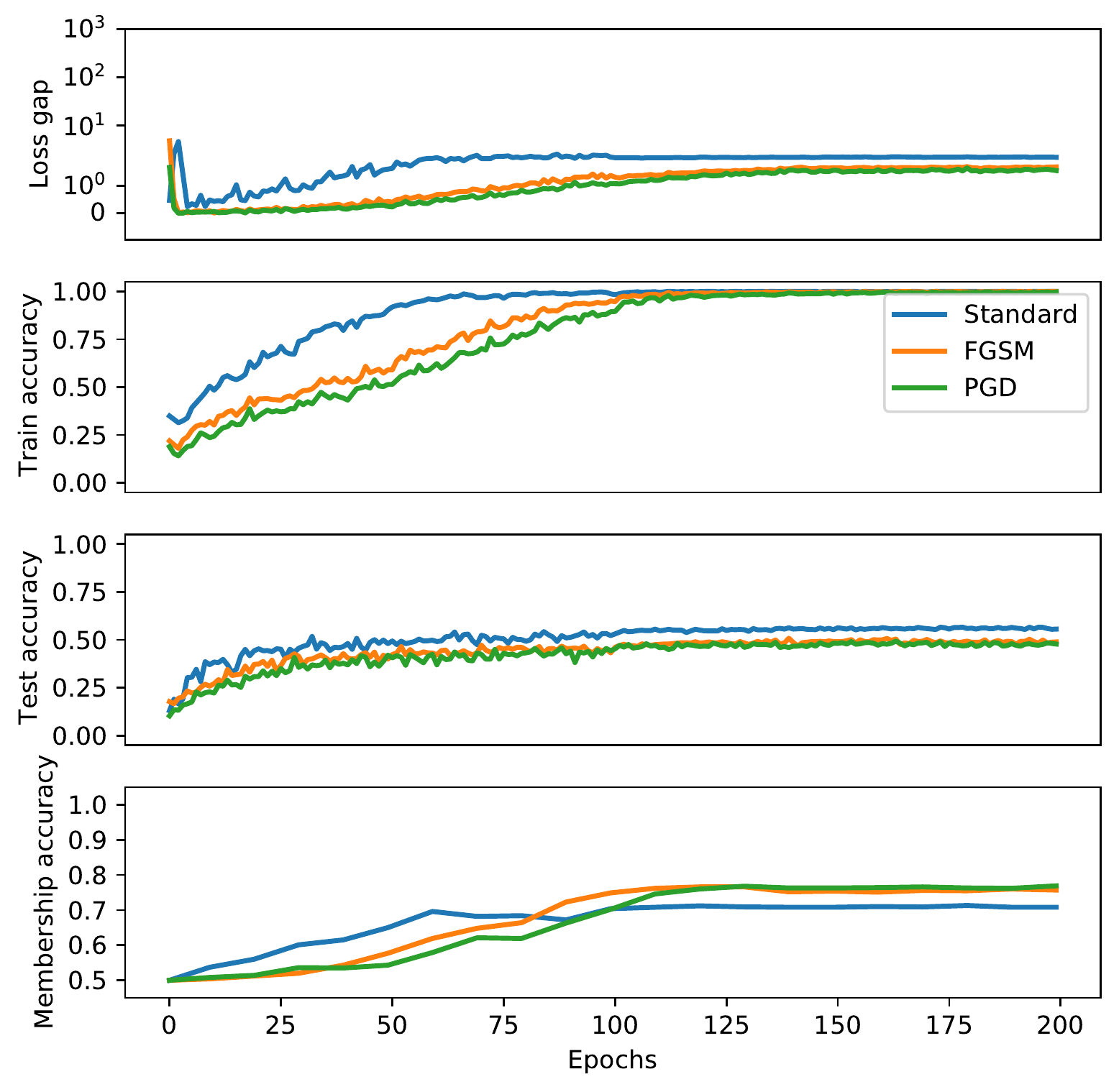}
  \caption{Training set size: 1000.}
  \label{fig:cifar10_mem_inf_train_vs_epoch_mean_transform_1000}
\end{subfigure}%
\begin{subfigure}{.33\textwidth}
  \centering
  \includegraphics[width=\linewidth]{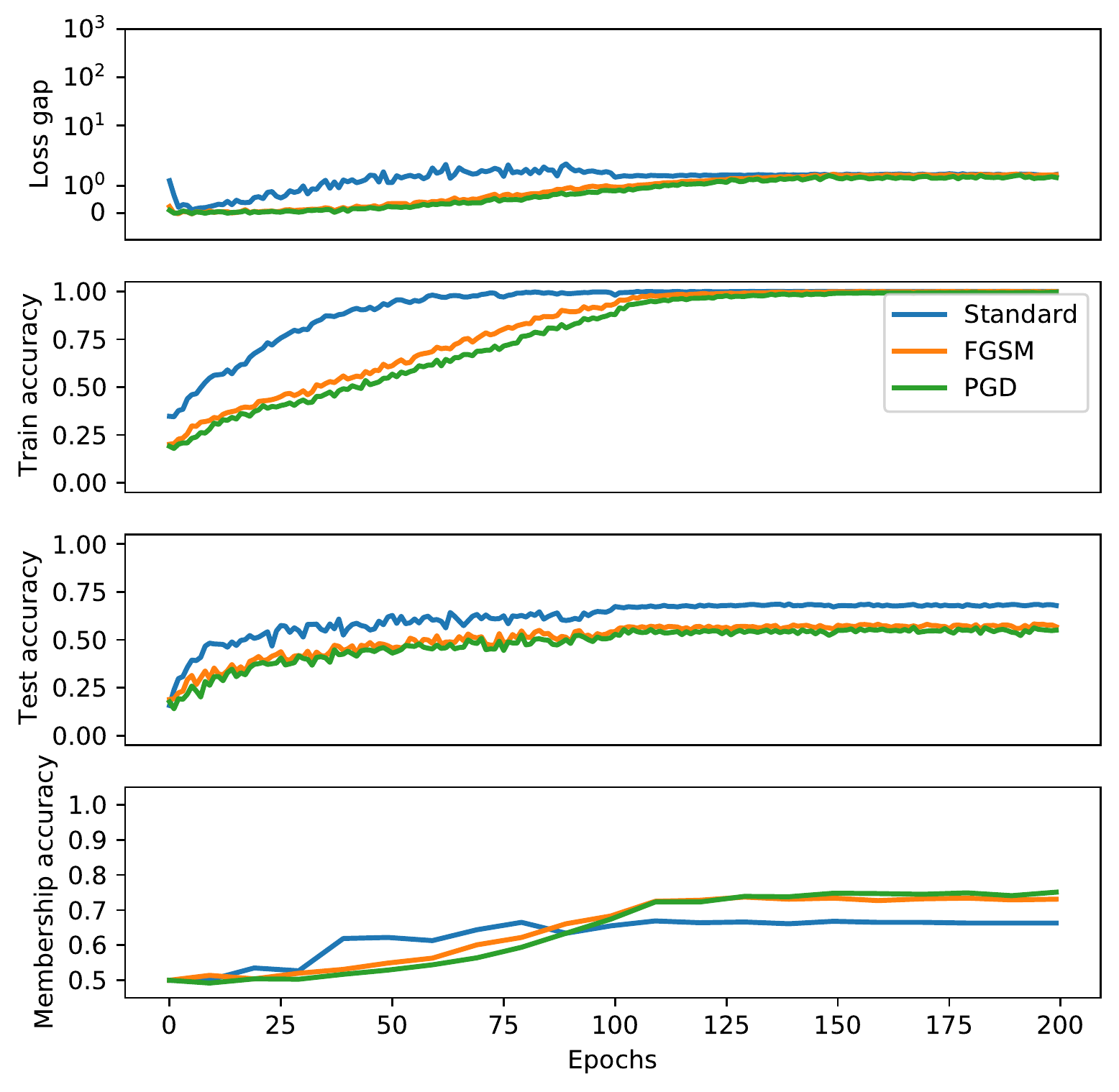}
  \caption{Training set size: 2000.}
  \label{fig:cifar10_mem_inf_train_vs_epoch_mean_transform_2000}
\end{subfigure}%
\begin{subfigure}{.33\textwidth}
  \centering
  \includegraphics[width=\linewidth]{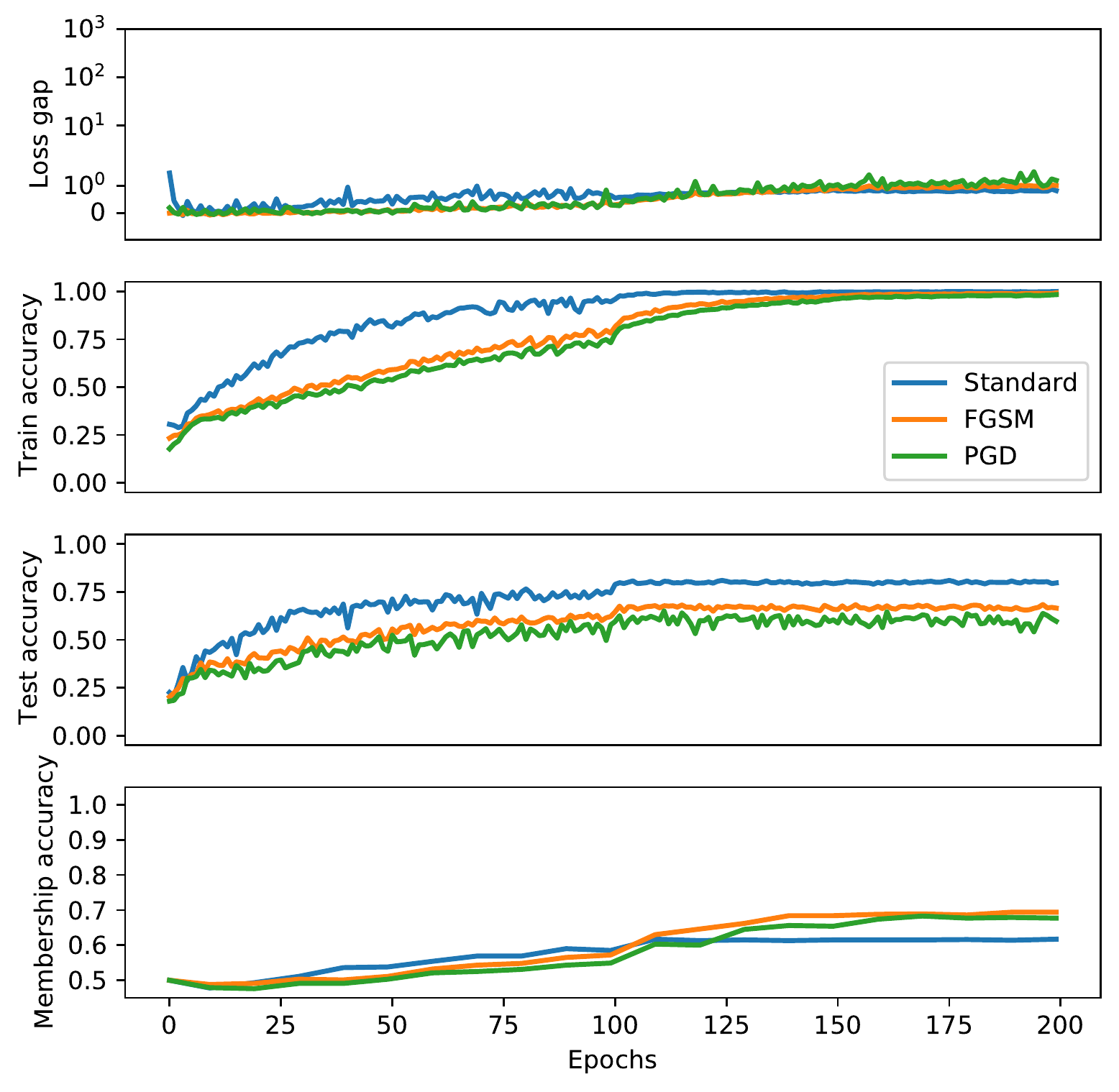}
  \caption{Training set size: 5000.}
  \label{fig:cifar10_mem_inf_train_vs_epoch_mean_transform_5000}
\end{subfigure}

\begin{subfigure}{.33\textwidth}
  \centering
  \includegraphics[width=\linewidth]{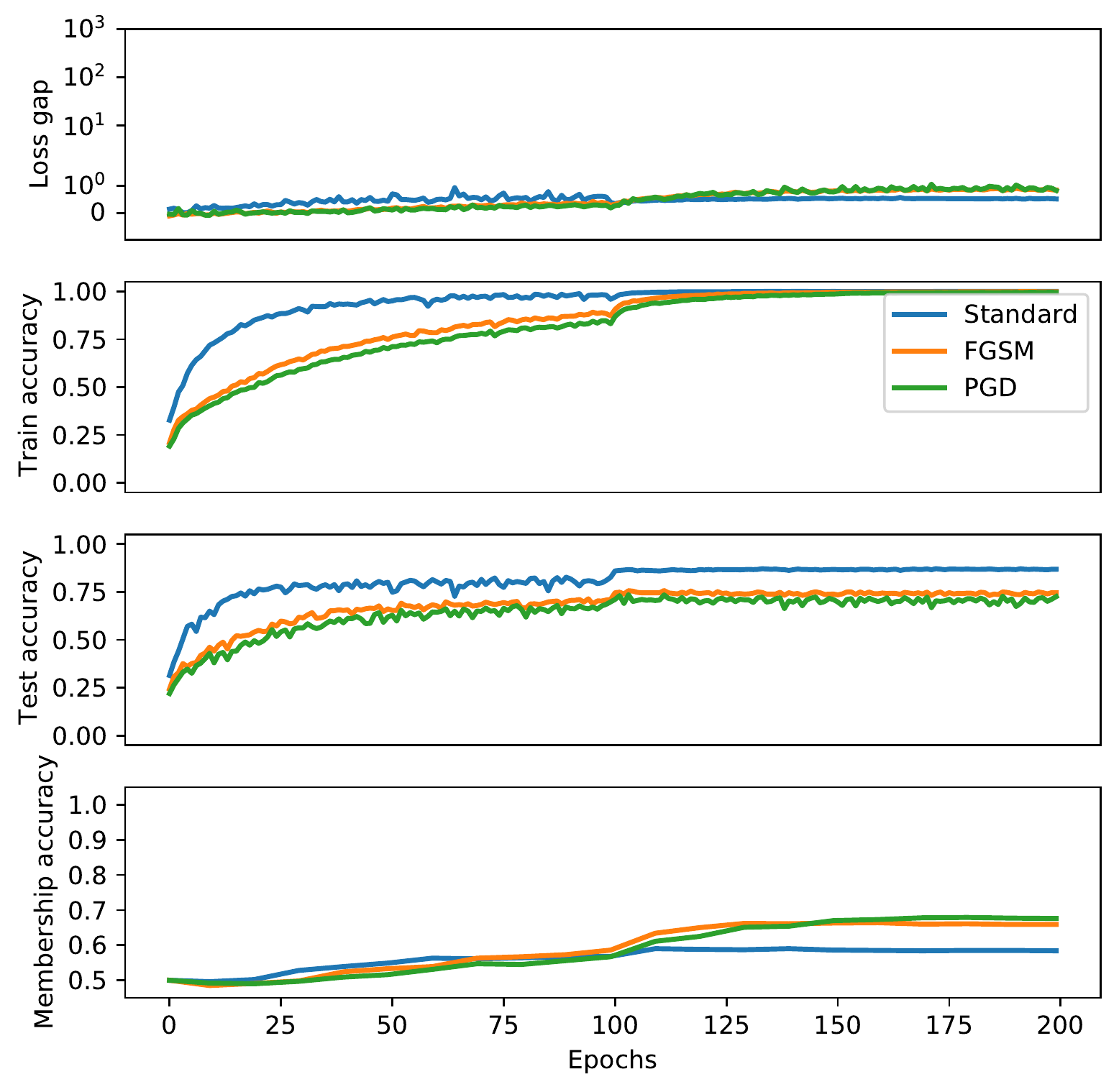}
  \caption{Training set size: 10000.}
  \label{fig:cifar10_mem_inf_train_vs_epoch_mean_transform_10000}
\end{subfigure}%
\begin{subfigure}{.33\textwidth}
  \centering
  \includegraphics[width=\linewidth]{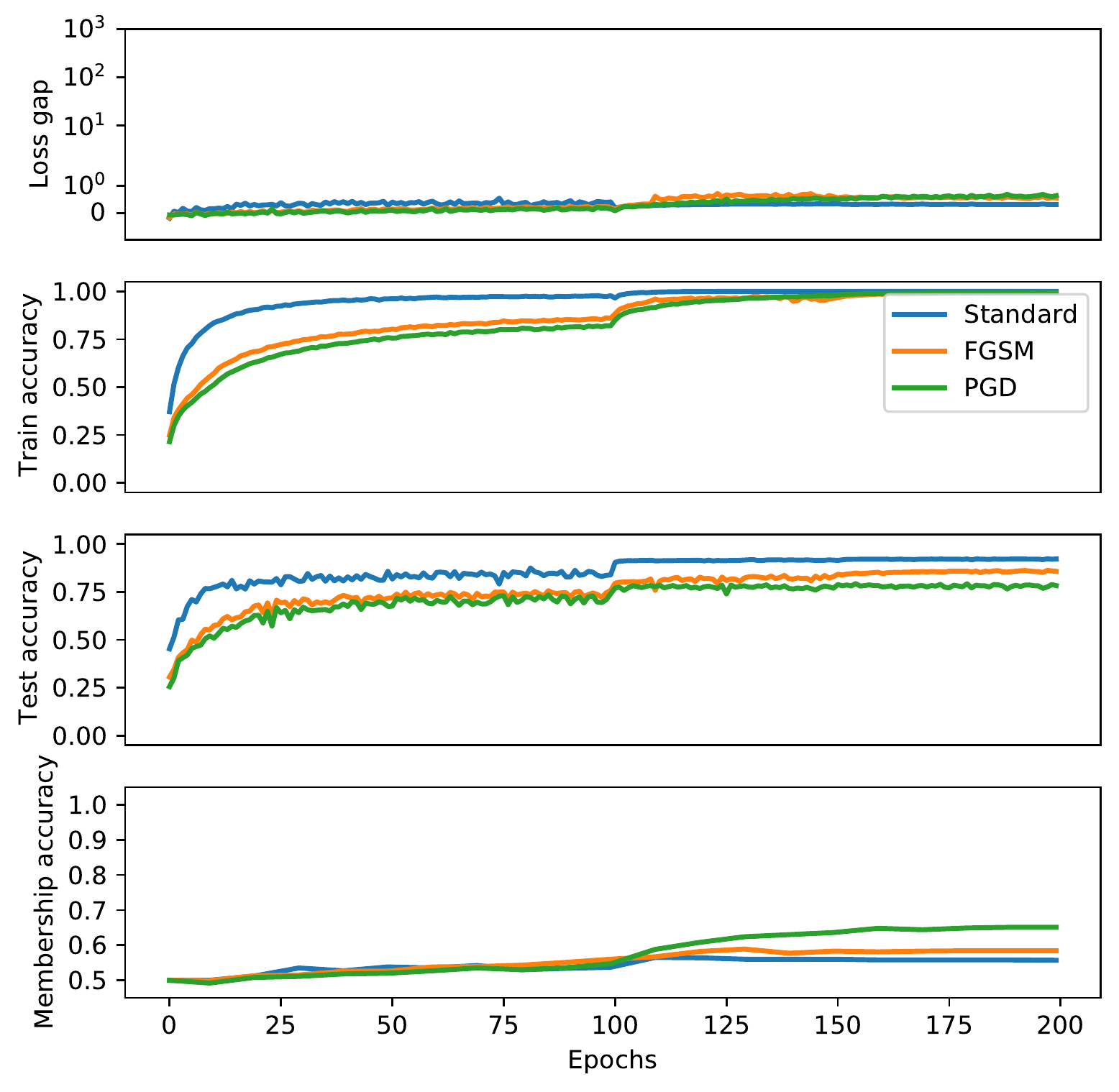}
  \caption{Training set size: 20000.}
  \label{fig:cifar10_mem_inf_train_vs_epoch_mean_transform_20000}
\end{subfigure}%
\begin{subfigure}{.33\textwidth}
  \centering
  \includegraphics[width=\linewidth]{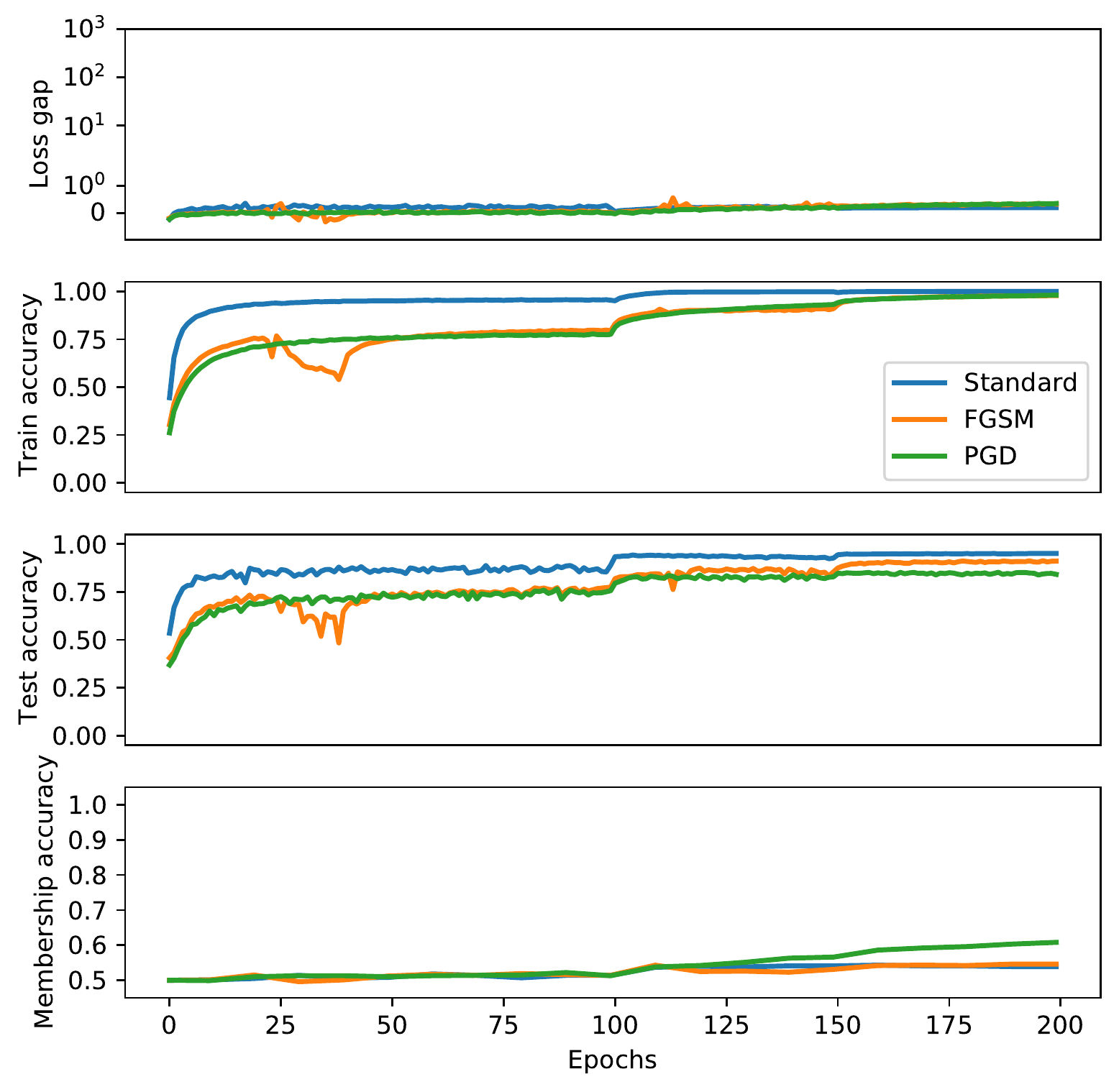}
  \caption{Training set size: 50000.}
  \label{fig:cifar10_mem_inf_train_vs_epoch_mean_transform_50000}
\end{subfigure}
\caption{Membership attack accuracy (using a mean threshold), loss gap, training and test accuracy throughout training on CIFAR-10, for standard and robust models.}
\label{fig:cifar10_mem_inf_train_vs_epoch_mean_transform}
\end{figure}

\begin{figure}
\centering
\begin{subfigure}{.33\textwidth}
  \centering
  \includegraphics[width=\linewidth]{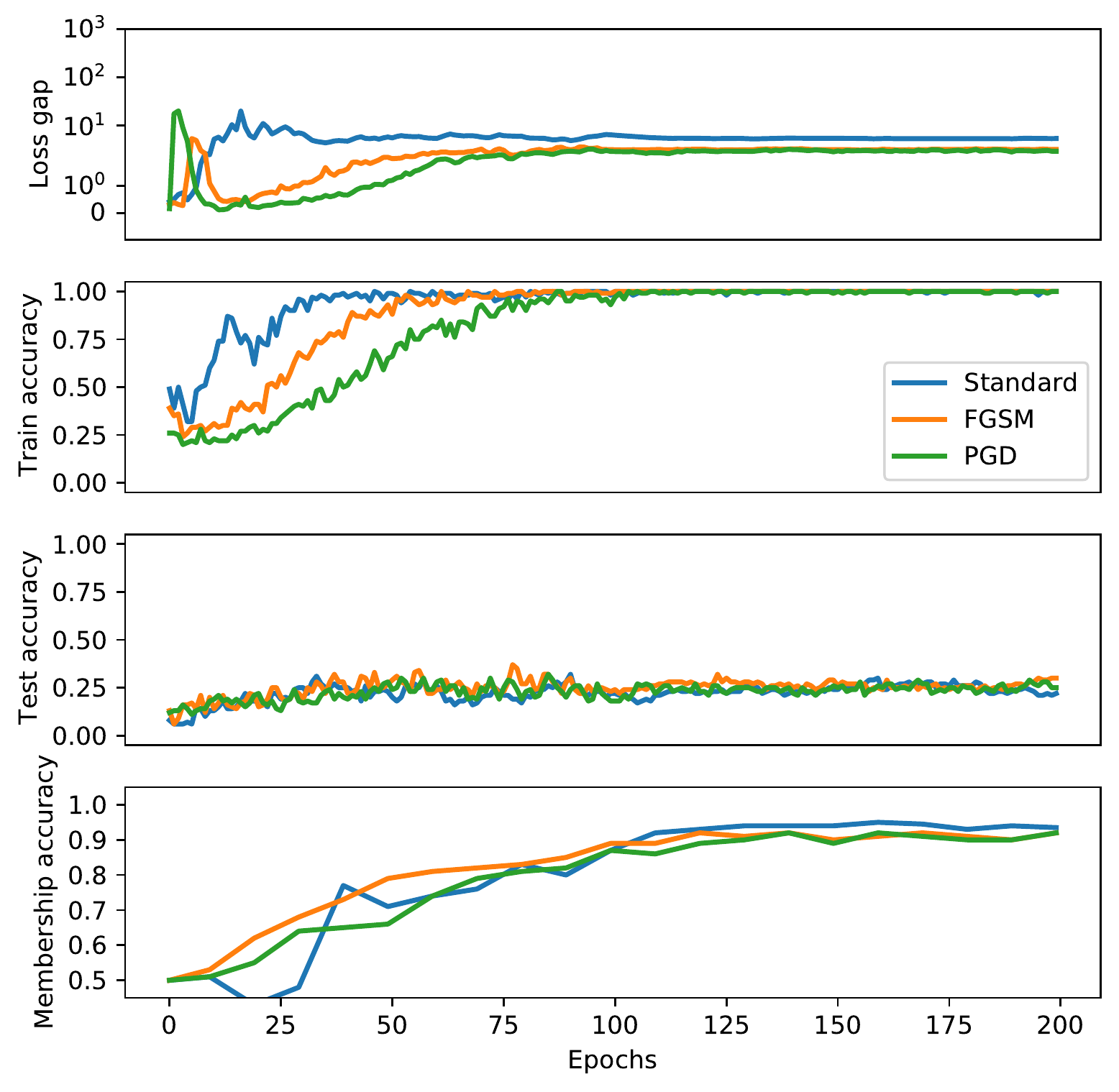}
  \caption{Training set size: 100.}
  \label{fig:cifar10_mem_inf_train_vs_epoch_median_transform_100}
\end{subfigure}%
\begin{subfigure}{.33\textwidth}
  \centering
  \includegraphics[width=\linewidth]{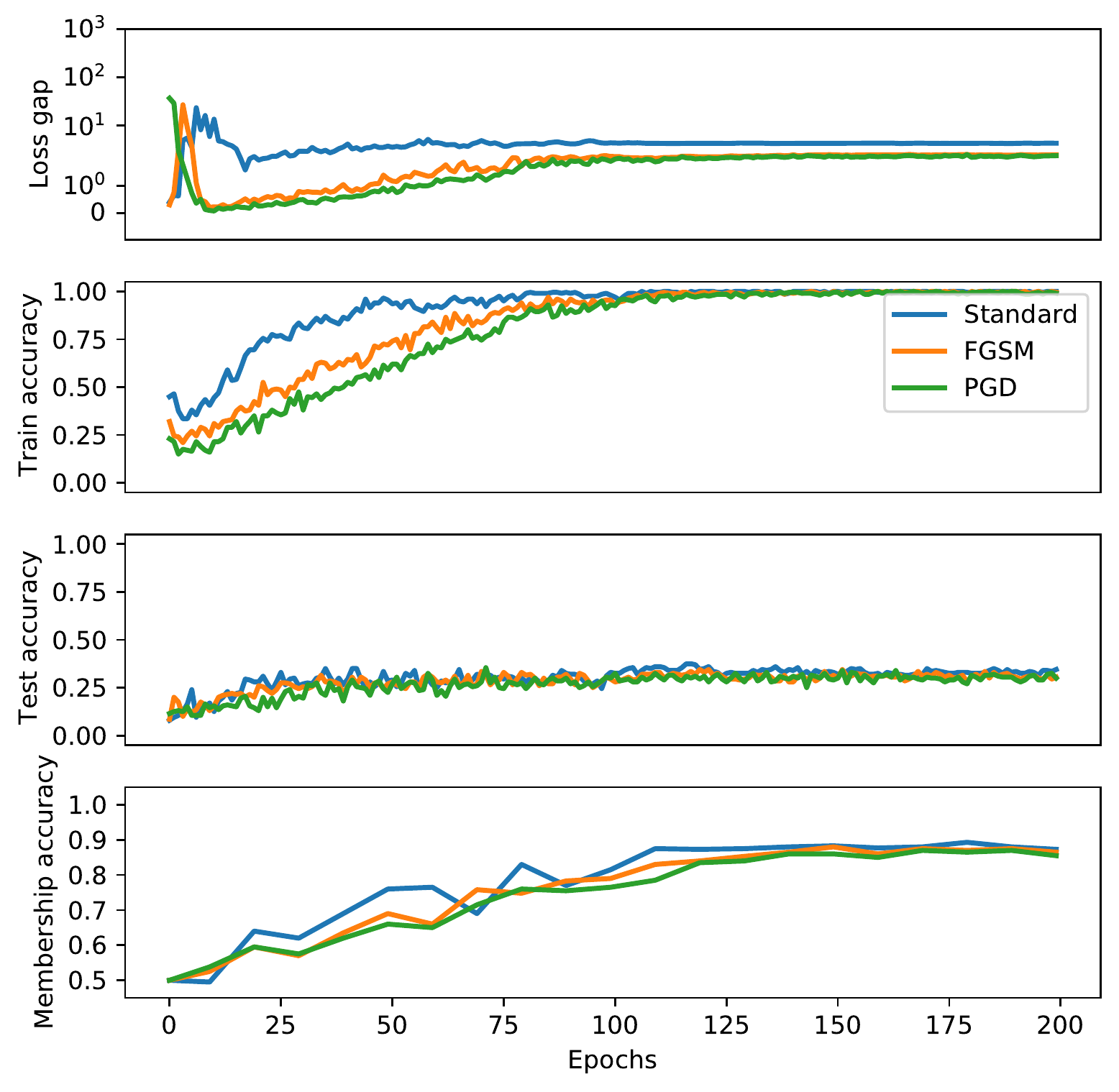}
  \caption{Training set size: 200.}
  \label{fig:cifar10_mem_inf_train_vs_epoch_median_transform_200}
\end{subfigure}%
\begin{subfigure}{.33\textwidth}
  \centering
  \includegraphics[width=\linewidth]{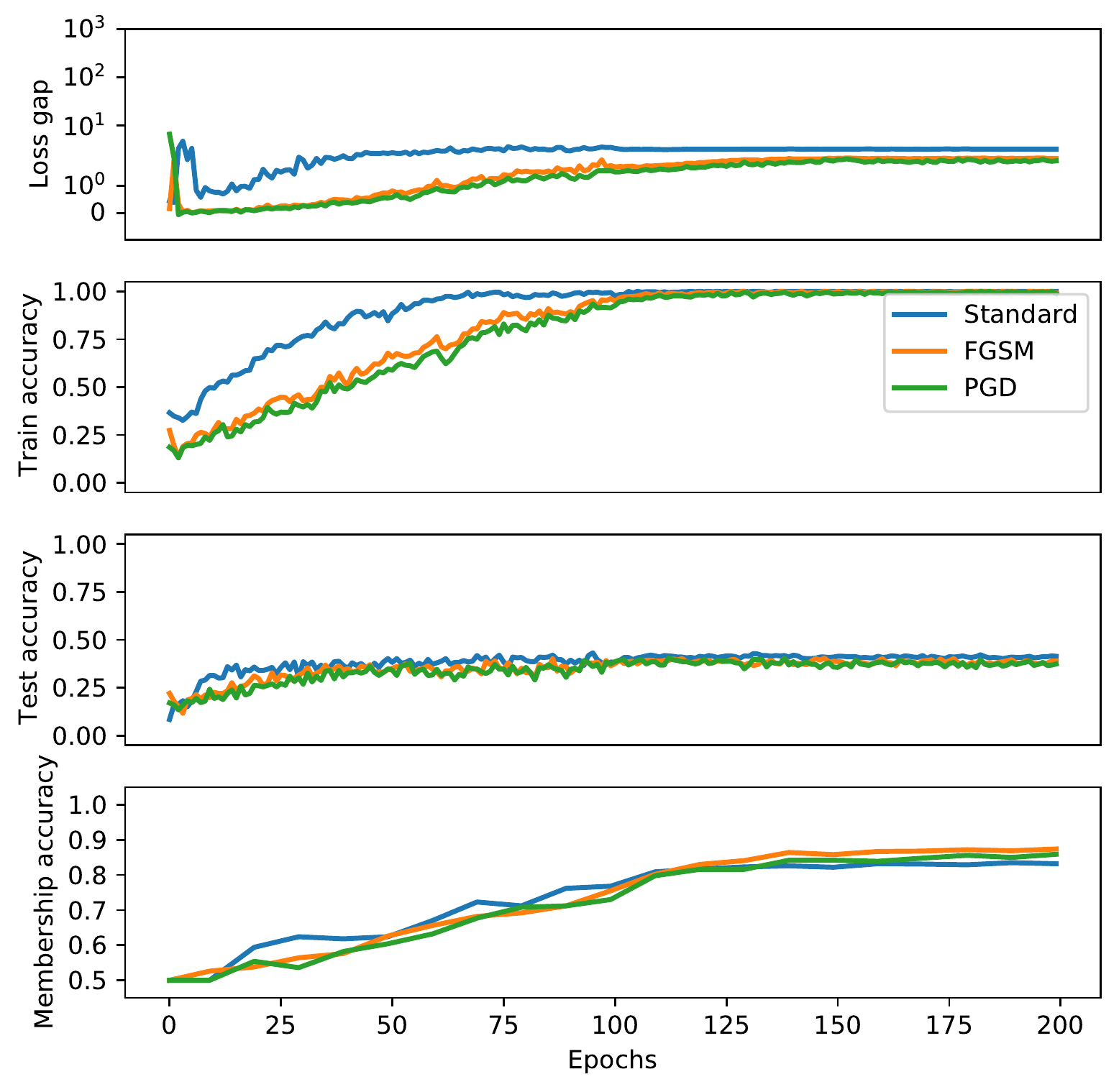}
  \caption{Training set size: 500.}
  \label{fig:cifar10_mem_inf_train_vs_epoch_median_transform_500}
\end{subfigure}

\begin{subfigure}{.33\textwidth}
  \centering
  \includegraphics[width=\linewidth]{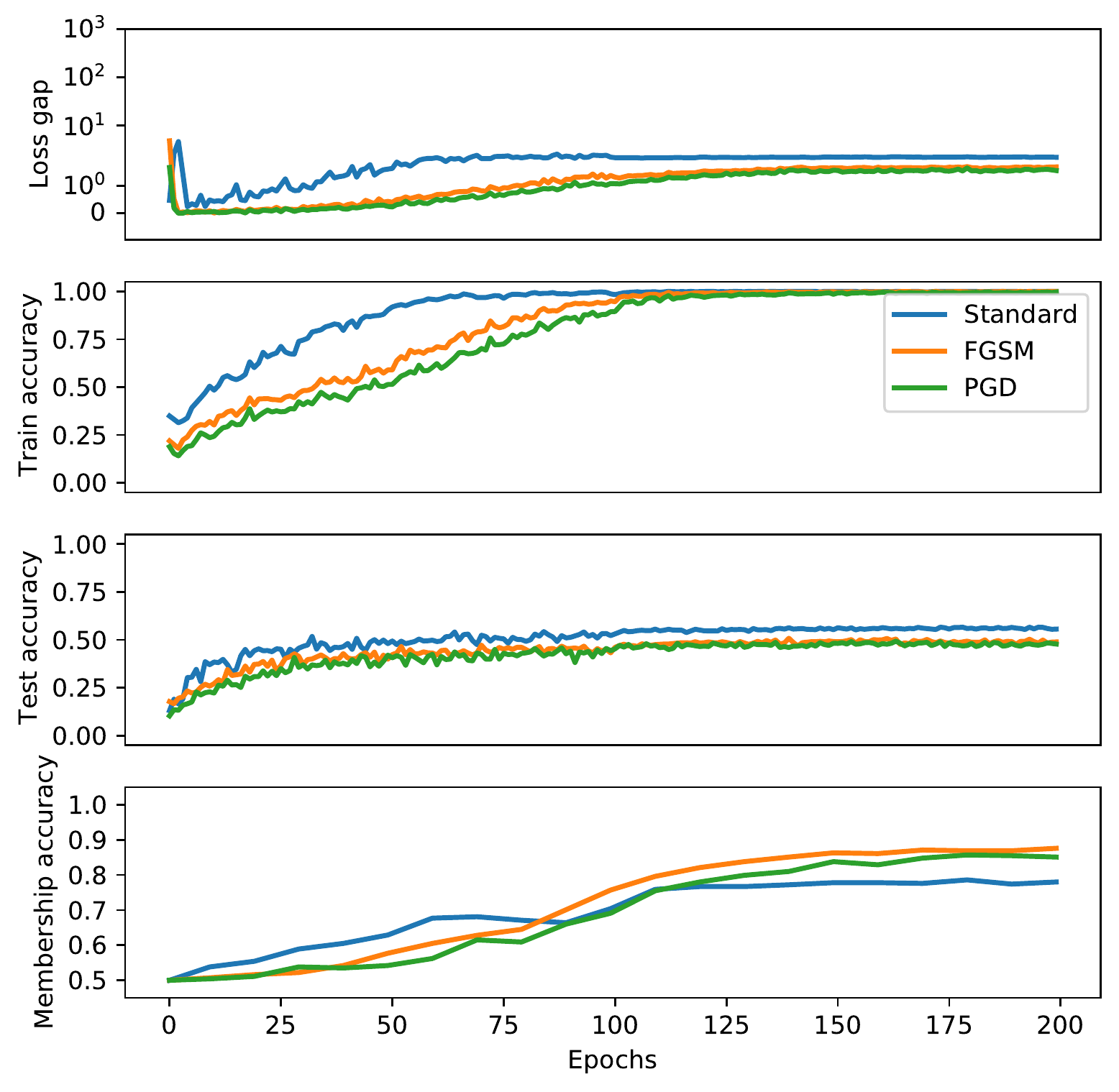}
  \caption{Training set size: 1000.}
  \label{fig:cifar10_mem_inf_train_vs_epoch_median_transform_1000}
\end{subfigure}%
\begin{subfigure}{.33\textwidth}
  \centering
  \includegraphics[width=\linewidth]{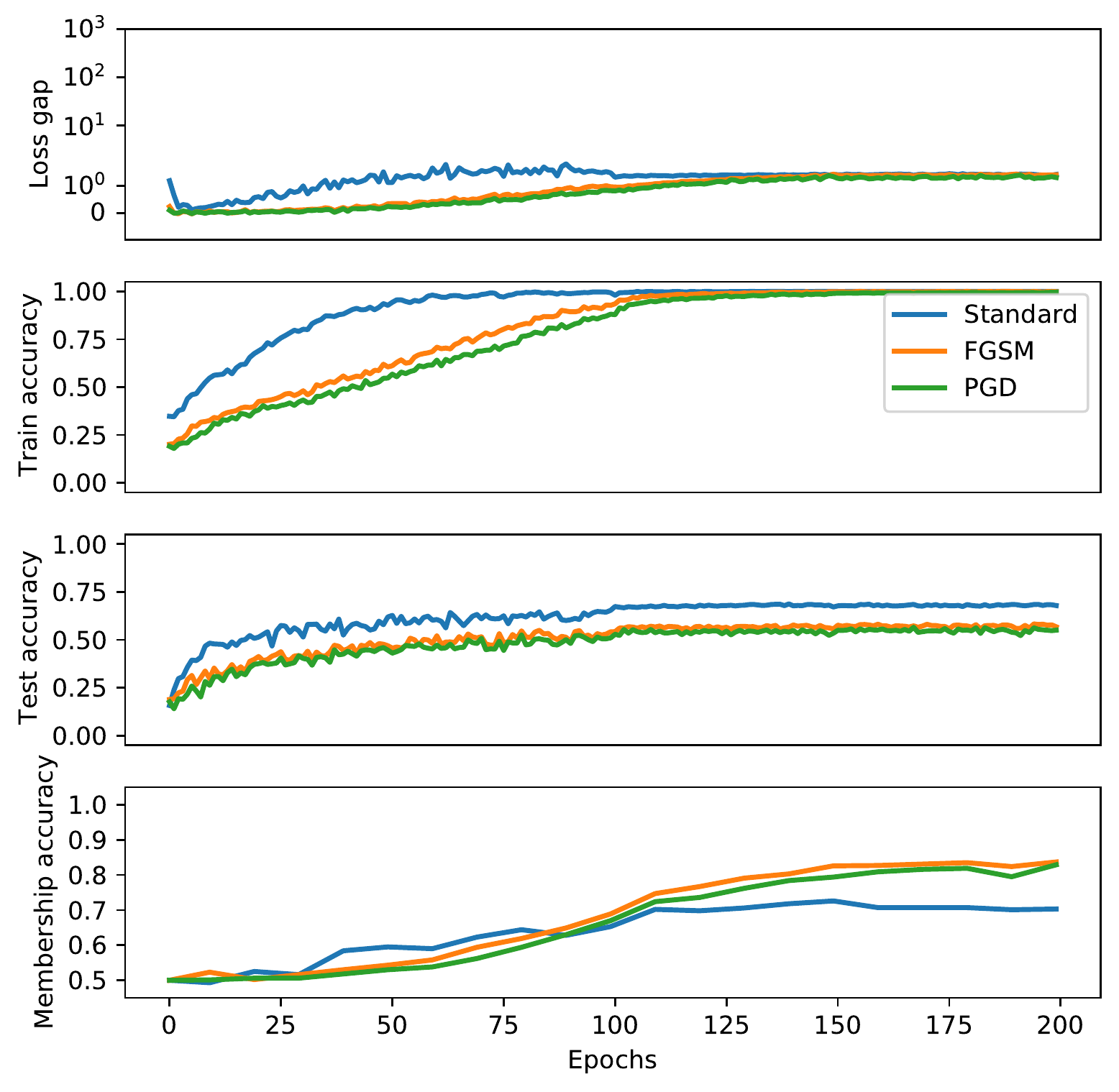}
  \caption{Training set size: 2000.}
  \label{fig:cifar10_mem_inf_train_vs_epoch_median_transform_2000}
\end{subfigure}%
\begin{subfigure}{.33\textwidth}
  \centering
  \includegraphics[width=\linewidth]{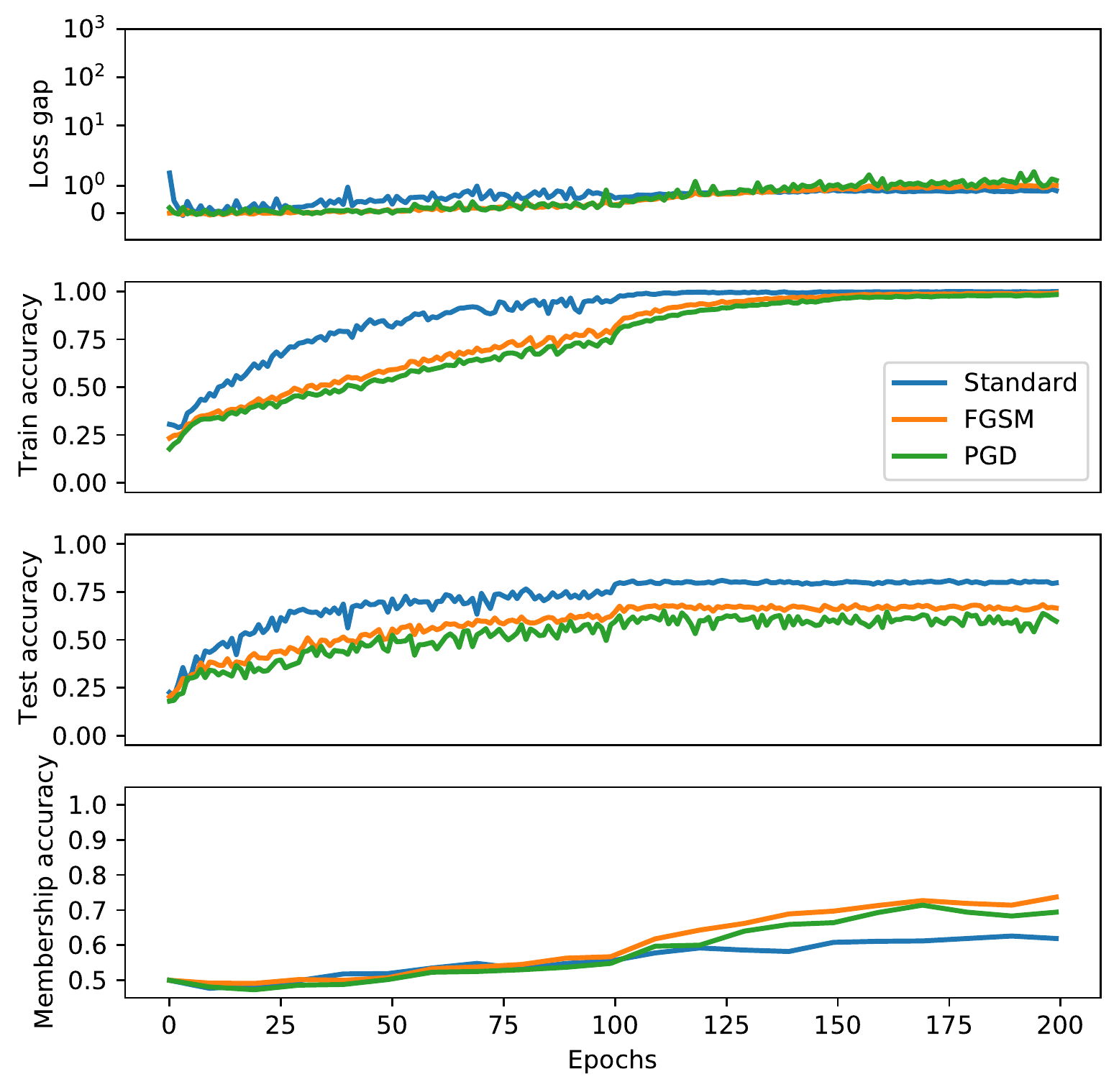}
  \caption{Training set size: 5000.}
  \label{fig:cifar10_mem_inf_train_vs_epoch_median_transform_5000}
\end{subfigure}

\begin{subfigure}{.33\textwidth}
  \centering
  \includegraphics[width=\linewidth]{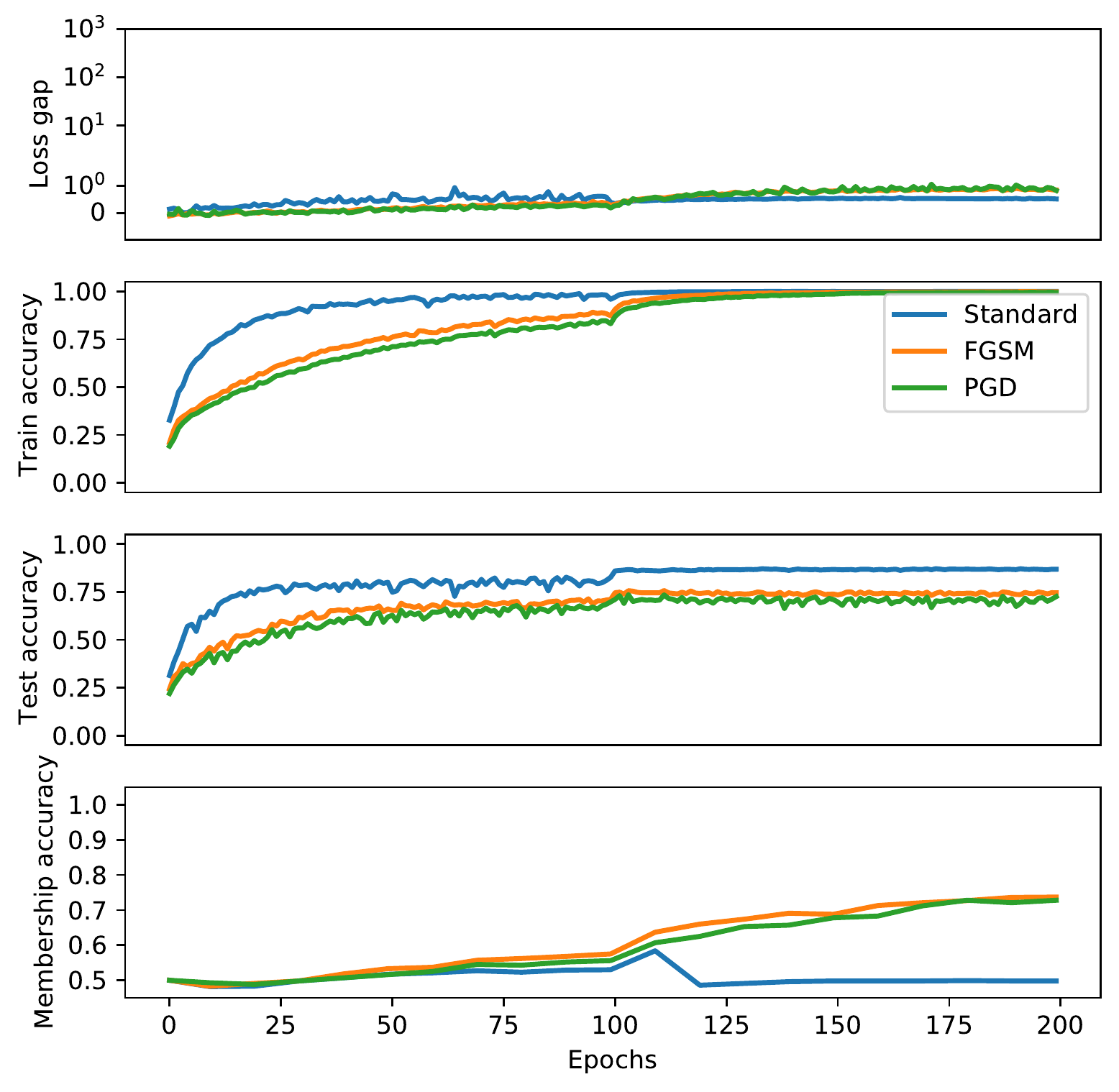}
  \caption{Training set size: 10000.}
  \label{fig:cifar10_mem_inf_train_vs_epoch_median_transform_10000}
\end{subfigure}%
\begin{subfigure}{.33\textwidth}
  \centering
  \includegraphics[width=\linewidth]{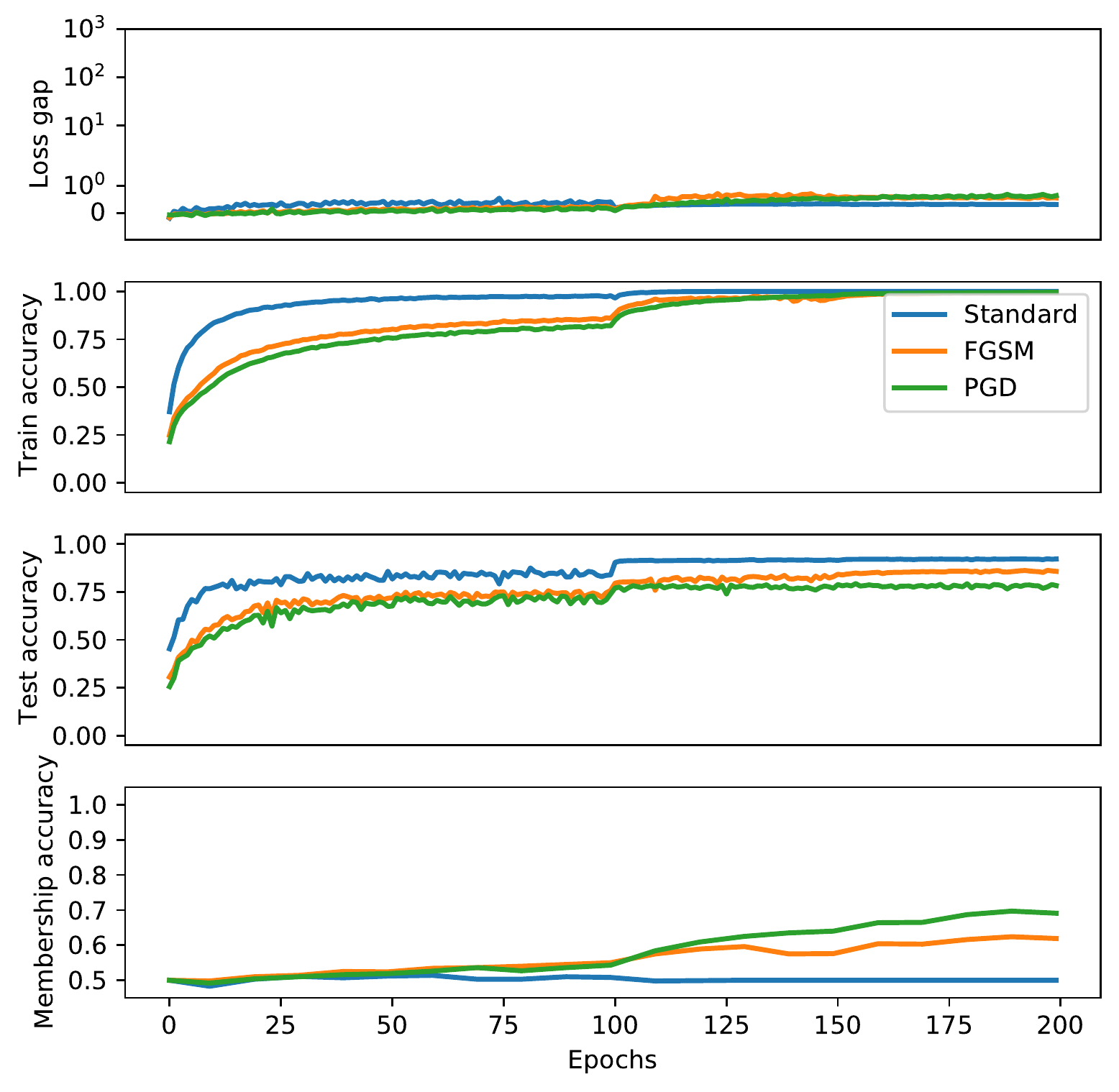}
  \caption{Training set size: 20000.}
  \label{fig:cifar10_mem_inf_train_vs_epoch_median_transform_20000}
\end{subfigure}%
\begin{subfigure}{.33\textwidth}
  \centering
  \includegraphics[width=\linewidth]{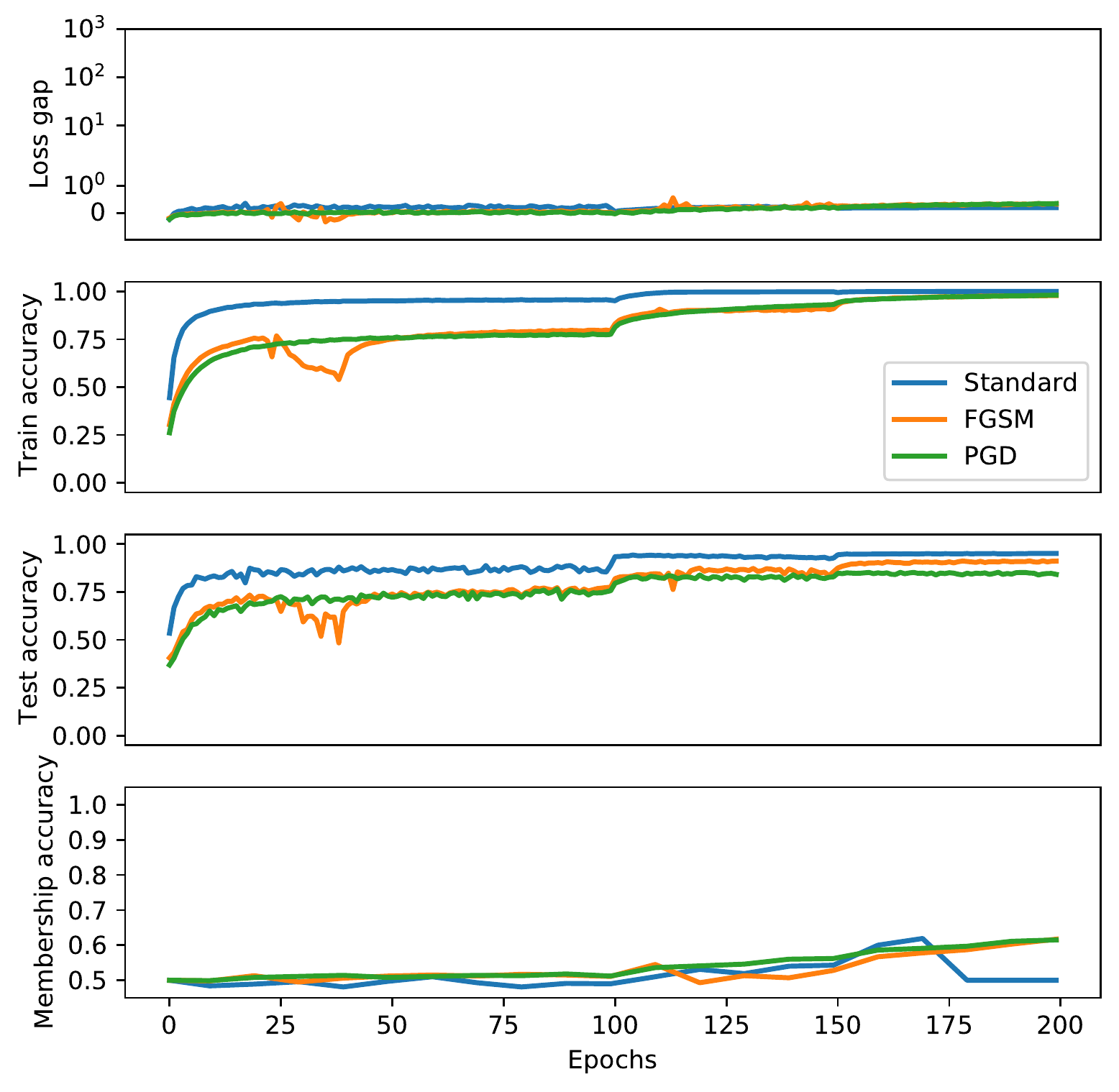}
  \caption{Training set size: 50000.}
  \label{fig:cifar10_mem_inf_train_vs_epoch_median_transform_50000}
\end{subfigure}
\caption{Membership attack accuracy (using a median threshold), loss gap, training and test accuracy throughout training on CIFAR-10, for standard and robust models.}
\label{fig:cifar10_mem_inf_train_vs_epoch_median_transform}
\end{figure}

Here, we plot the membership accuracy, the loss gap, training accuracy, and test accuracy throughout training for each training set size under the experiments defined in \cref{sec:cifar10_results} on the CIFAR-10 dataset.
\Cref{fig:cifar10_mem_inf_train_vs_epoch_mean_transform} shows results for a membership attack using a mean loss threshold, and \cref{fig:cifar10_mem_inf_train_vs_epoch_median_transform} results for a membership attack using a median loss threshold.

\newpage

\begin{figure*}[t]
  \centering
  \includegraphics[width=0.5\linewidth]{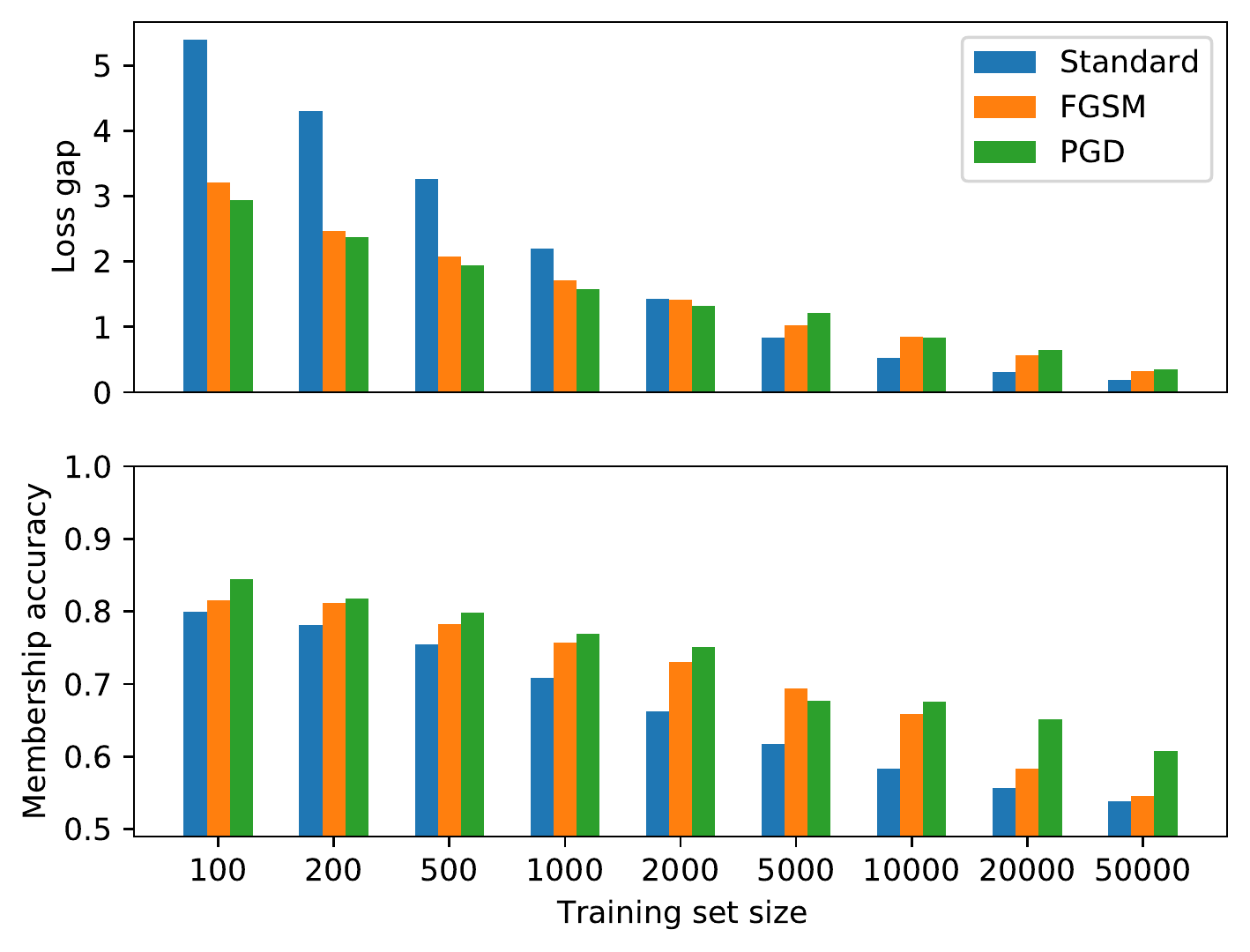}
\caption{Membership attack accuracy and loss gap on CIFAR-10 as a function of the training set size, for standard and robust models (using adversarial training with $\epsilon=\frac{8}{255}$ and using either the FGSM or PGD attack). We use the MALT membership attack~\citep{sablayrolles2019white} with a mean decision threshold as explained in~\cref{sec:cifar10_results}.}
\label{fig:cifar10_mem_inf_train_vs_size_mean_threshold}
\end{figure*}

In \cref{fig:cifar10_mem_inf_train_vs_size_mean_threshold}, we plot the membership accuracy and loss gap for the MALT attack with a mean threshold as described in \cref{sec:cifar10_results}. 
The membership accuracy using the mean threshold is strictly worse than the median threshold.

\clearpage
\section{\citet{chen2020more} generalization bounds}
\label{sec:chen_gen_bounds}

In this section, we give an improved bound for the number of training inputs to ensure the generalization gap as defined in \citet{chen2020more} is increasing. 
Following this, we study how the generalization gap is affected by label corruption.

\subsection{An improved upper bound}
\label{ssec:improved_lower_bounds}

Here, we give an improved bound to~\citet{chen2020more} for the number of training inputs, $n$, required to ensure the generalization gap, $g_n = \mathbb{E}_{(x,y)\sim\mathcal{D}}[y\inp{\theta^{\text{std}} - \theta^{\text{rob}}}{x}]$, increases with $n$.

We first note that this value is governed by the function $2\Phi(\frac{\sqrt{n}\mu}{\sigma})-\Phi(\frac{\sqrt{n}\mu}{\sigma}(1+\frac{\epsilon}{\mu})) -\Phi(\frac{\sqrt{n}\mu}{\sigma}(1-\frac{\epsilon}{\mu}))$, which can be written as

\begin{align}
\kappa(x) 
&=  2\Phi(x)-\Phi(x(1+\delta)) -\Phi(x(1-\delta)) 
\end{align}

where $x=\frac{\sqrt{n}\mu}{\sigma}$ and $\delta=\frac{\epsilon}{\mu}$ -- see \citet{chen2020more} for further details.
To find if $\kappa(x)$ is increasing or decreasing we look at:

\begin{align}
    \kappa'(x) = \frac{1}{\sqrt{2\pi}}e^{\frac{1}{2}(1+\delta)^2x^2}(2e^{\frac{1}{2}\delta(2+\delta)x^2} + (\delta-1)e^{2\delta x^2} - 1 - \delta)
\end{align}

Let $\rho(x) = 2e^{\frac{1}{2}\delta(2+\delta)x^2} + (\delta-1)e^{2\delta x^2} - 1 - \delta$, then $\kappa'(x)=0 \iff x=\pm\infty$ or $\rho(x)=0$.

Now, $\rho(x)\geq \nu(x)=(\delta-1)e^{2\delta x^2}+ 2e^{\delta x^2} - 1 - \delta$.
Let $y=e^{x^2}$, then:

\begin{align}
    &\nu(x) = 0 \\
    \iff &(\delta - 1)y^{2\delta} + 2y^{\delta} = 1+\delta \\
    \iff &y=1 \text{ or } y=(\frac{1+\delta}{1-\delta})^{\frac{1}{\delta}}.
\end{align}

Now, $y=1 \implies x=0$ and $y=(\frac{1+\delta}{1-\delta})^{\frac{1}{\delta}} \implies x^2 = \frac{1}{\delta}\log(\frac{1+\delta}{1-\delta})$.
Observe that $x^2$ achieves a global minimum as $\delta\rightarrow 0^+$, so:

\begin{align}
    \lim_{\delta\rightarrow 0^+} \frac{1}{\delta}\log(\frac{1+\delta}{1-\delta}) = \lim_{\delta\rightarrow 0^+} \frac{2}{1-\delta^2} = 2
\end{align}

Hence $x^2\geq 2 > \frac{3}{2}$, $\forall \delta >0$.

We also have:

\begin{align}
    \frac{1}{\delta}\log(\frac{1+\delta}{1-\delta}) > 2\log(\frac{1}{1-\delta})
\end{align}

holds for $0<\delta<0.71$. Thus for $\delta\in(0,1)$ there exists


\begin{align}
    \sqrt{\max(\frac{1}{\delta}\log(\frac{1+\delta}{1-\delta}),  2\log(\frac{1}{1-\delta}))} < x_0
\end{align}

such that $\kappa(x)$ is stricly increasing on $(0,x_0)$. It follows that $g_n$ is strictly increasing when

\begin{align}
    n \leq \min_{j\in[d], \mu_j>0}\max \bigg(\frac{\mu_j}{\epsilon}\log(\frac{\mu_j +\epsilon}{\mu_j -\epsilon}), 2\log(\frac{\mu_j}{\mu_j -\epsilon})\bigg)\big(\frac{\sigma_j}{\mu_j}\big)^2 \label{eq:better_chen_bound}
\end{align}

\Cref{eq:better_chen_bound} gives a tighter upper bound than the \citet{chen2020more} bound given by

\begin{align}
    n \leq \min_{j\in[d], \mu_j>0}\max \bigg(\frac{3}{2}, 2\log(\frac{\mu_j}{\mu_j -\epsilon})\bigg)\big(\frac{\sigma_j}{\mu_j}\big)^2 
\end{align}

\subsection{\citet{chen2020more} generalization bounds under label noise}
\label{ssec:label_noise}

Let $y\sim\{\pm1\}$ uniformly at random, and $x\in\mathbb{R}^d$, where $\forall j\in[d]$,

\begin{align}
    x_j = \begin{cases}
    \sim\mathcal{N}(y\mu_j,\sigma_j^2) , & \text{w.p } \zeta \\
    \sim\mathcal{N}(-y\mu_j,\sigma_j^2) , & \text{w.p } 1-\zeta \\
  \end{cases}
\end{align}

and $\nicefrac{1}{2}<\zeta\leq1$.
Then under \citet{chen2020more} the generalization gap is given by:

\begin{align}
    g_n &= \mathbb{E}_{(x,y)\sim\mathcal{D}}[y\inp{\theta^{\text{std}} - \theta^{\text{rob}}}{x}] \\
    &= \inp{\theta^{\text{std}} - \theta^{\text{rob}}}{(2\zeta -1)\mu}
\end{align}

Following theorem 1 in \citet{chen2020more}, $g_n$ is strictly increasing when:

\begin{align}
    n \leq \min_{j\in[d], \mu_j>0}\max \bigg(\frac{2\zeta-1}{\epsilon}\mu_j\log(\frac{(2\zeta-1)\mu_j +\epsilon}{(2\zeta-1)\mu_j -\epsilon}), 2\log(\frac{(2\zeta-1)\mu_j}{(2\zeta-1)\mu_j -\epsilon})\bigg)\big(\frac{\sigma_j}{(2\zeta-1)\mu_j}\big)^2
\end{align}

Thus the upper bound increase as $\sim\mathcal{O}(\frac{1}{\zeta^2})$, providing other hyperparameters are fixed.

\section{Bayes risk analysis}
\label{sec:bayes_risk_analysis}

In this section, we show that a linear classifier with a linear loss can learn to separate Gaussian data as defined \cref{sec:gaussian_membership}, with an error rate equal to the Bayes error.
Let $F_{\theta}(x)=\sign(f_{\theta}(x))$ where $f_{\theta}(x) = \inp{\theta}{x}$ and $\ell =-y\inp{\theta}{x}$.

From~\cref{sec:gaussian_membership}, $\theta_n^{\text{std}}=\gamma\sign(u)$ and in the infinite data limit $\sign(u) = \sign(\mu)=1$, and so $\theta^{\text{std}} = \gamma >0$.
Similarly $\theta_n^{\text{rob}}=\gamma\sign(u-\epsilon\sign(u))$, and in the infinite data limit $\theta^{\text{rob}} = \gamma\sign(\mu - \epsilon) > 0$ if $\epsilon < \mu$. 
Next, we show that any linear classifier with $\theta > 0$ is equivalent to the Bayes classifier in error rate.

Consider the Gaussian class-conditional densities:

\begin{align}
    &P(x\mid y=1) = \frac{1}{(2\pi)^{\frac{d}{2}}\sigma}\exp{(-\frac{1}{2}(x-\bar{\mu})^{T}\sigma^2I(x-\bar{\mu}) )}\\
    &P(x\mid y=-1) = \frac{1}{(2\pi)^{\frac{d}{2}}\sigma}\exp{(-\frac{1}{2}(x+\bar{\mu})^{T}\sigma^2 I(x+\bar{\mu}) )}\\
\end{align}

where $\bar{\mu}^T=[\mu, \dots, \mu]$ is a $d$-dimensional vector. 
The Bayes decision rule is given by $\frac{2\bar{\mu}^{T}}{\sigma^2}x \gtrless 0$. 
Note that $\frac{2\bar{\mu}^{T}}{\sigma^2}x > 0 \iff x>0$, since $\frac{2\bar{\mu}^{T}}{\sigma^2}>0$. 
Clearly then, we have a decision rule that is optimal if the sign of $x$ is not flipped. 
That is, for $\theta\in\mathbb{R}^d$, $\sign(\sum_{j=1}^d \theta_jx_j) = \sign(\sum_{j=1}^d x_j)$.

Note, if $x_j\sim \mathcal{N}(y\mu, \sigma^2)$, $\forall j\in [d]$,
then $\sum_{j=1}^d x_j \sim \mathcal{N}(yd\mu, d\sigma^2)$.
Without loss of generality, let $y=1$, then $P(\sum_{i=1}^d x_i >0)=\Phi(\frac{\sqrt{d}\mu}{\sigma})$. 
As $d\rightarrow \infty$, $\Phi(\frac{\sqrt{d}\mu}{\sigma})\rightarrow 1$, and so $P(\sum_{i=1}^d \theta_ix_i > 0)\rightarrow1$ if $\theta > 0$. 
Clearly then any $\theta >0$ gives the Bayes optimal classifier in $d$ dimensions and the Bayes error approaches zero as $d$ increases.

So both $\theta^{\text{std}} = \gamma$ and $\theta^{\text{rob}} = \gamma\sign(\mu - \epsilon)$ if $\epsilon < \mu$, are classifiers with the Bayes error rate.

\end{document}